\documentclass[10pt,journal,compsoc]{IEEEtran}
\ifCLASSOPTIONcompsoc
  \usepackage[nocompress]{cite}
\else
  \usepackage{cite}
\fi
\ifCLASSINFOpdf
\else
\fi
\newcommand{\ie}{\textit{i}.\textit{e}.}
\newcommand{\eg}{\textit{e}.\textit{g}.}
\usepackage[dvipsnames]{xcolor}
\usepackage{graphicx}
\usepackage{enumerate}
\usepackage{amsmath,amssymb}
\usepackage{multirow}
\usepackage{verbatim}
\usepackage{threeparttable}
\usepackage{subfigure}
\usepackage[colorlinks]{hyperref}
\usepackage{color, colortbl}

\begin{document}
%
\title{Affinity Attention Graph Neural Network for Weakly Supervised Semantic Segmentation}
%
%
%
%

\author{Bingfeng Zhang, 
	Jimin Xiao,~\IEEEmembership{Member,~IEEE,}
	Jianbo Jiao, ~\IEEEmembership{Member,~IEEE,} \\
	Yunchao Wei,~\IEEEmembership{Member,~IEEE,} 
	Yao Zhao, ~\IEEEmembership{Senior Member,~IEEE} 
	\thanks{B.~Zhang is with University of Liverpool, UK, and also with School of Advanced Technology, Xi'an Jiaotong-Liverpool University, Suzhou, P.R. China  (e-mail: bingfeng.zhang@liverpool.ac.uk).}
	\thanks{J.~Xiao is with School of Advanced Technology, Xi'an Jiaotong-Liverpool University, Suzhou, P.R. China (e-mail: jimin.xiao@xjtlu.edu.cn). (Corresponding author: Jimin Xiao).}
	\thanks{J.~Jiao is with the Department of Engineering Science, University of Oxford, UK (e-mail: jianbo.jiao@eng.ox.ac.uk).}
	\thanks{Y.~Wei is with University of Technology Sydney, Australia (e-mail: yunchao.wei@uts.edu.au).}
	\thanks{Y. Zhao is with Institute of Information Science, Beijing Jiaotong University, Beijing, China (e-mail: yzhao@bjtu.edu.cn).}
}

%
%

\markboth{Submitted to IEEE TPAMI}%
{Shell \MakeLowercase{\textit{et al.}}: Bare Demo of IEEEtran.cls for Computer Society Journals}
%



\IEEEtitleabstractindextext{%
\begin{abstract}
Weakly supervised semantic segmentation is receiving great attention due to its low human annotation cost. In this paper, we aim to tackle bounding box supervised semantic segmentation, \ie, training accurate semantic segmentation models using bounding box annotations as supervision. To this end, we propose Affinity Attention Graph Neural Network (\emph{$A^2$GNN}). Following previous practices, we first generate pseudo semantic-aware seeds, which are then formed into semantic graphs based on our newly proposed affinity Convolutional Neural Network (CNN). Then the built graphs are input to our \emph{$A^2$GNN}, in which an affinity attention layer is designed to acquire the short- and long- distance information from soft graph edges to accurately propagate semantic labels from the confident seeds to the unlabeled pixels. However, to guarantee the precision of the seeds, 
we only adopt a limited number of confident pixel seed labels for \emph{$A^2$GNN}, which may lead to insufficient supervision for training. To alleviate this issue, we further introduce a new loss function and a consistency-checking mechanism to leverage the bounding box constraint, so that more reliable guidance can be included for the model optimization. Experiments show that our approach achieves new state-of-the-art performances on Pascal VOC 2012 datasets (\emph{val}: 76.5\%, \emph{test}: 75.2\%).  More importantly, our approach can be readily applied to bounding box supervised instance segmentation task or other weakly supervised semantic segmentation tasks, with state-of-the-art or comparable performance among almot all weakly supervised tasks on PASCAL VOC or COCO dataset. Our source code will be available at \href{https://github.com/zbf1991/A2GNN}{https://github.com/zbf1991/A2GNN}.

\end{abstract}

\begin{IEEEkeywords}
Weakly supervised, semantic segmentation, graph neural network.
\end{IEEEkeywords}}

\maketitle

\IEEEdisplaynontitleabstractindextext

%
\IEEEpeerreviewmaketitle

\section{Introduction}
%
%
%
%
\IEEEPARstart{W}{eakly} supervised semantic segmentation aims to make a pixel-level semantic prediction using weak annotations as supervision. According to the level of provided annotations, the weak supervision can be divided into scribble level~\cite{tang2018normalized,lin2016scribblesup,tang2018regularized}, bounding box level~\cite{dai2015boxsup,khoreva2017simple,song2019box,hsu2019weakly}, point level~\cite{bearman2016s} and image level~\cite{ahn2018learning,wei2016stc,hou2018self,ahn2019weakly}. In this paper, we mainly focus on bounding box supervised semantic segmentation (BSSS). The key challenge of BSSS lies in how to accurately estimate the pseudo object mask within the given bounding box so that reliable segmentation networks can be learned with the generated pseudo masks using current popular fully convolutional networks (FCN)~\cite{long2015fully,chen2017deeplab,chen2017rethinking, he2017mask}. 

Most previous practices~\cite{dai2015boxsup,khoreva2017simple,song2019box, papandreou1502weakly} for the BSSS task use object proposals~\cite{arbelaez2014multiscale,martin2001database} to provide some seed labels as supervision. These methods follow a common pipeline of employing object proposals~\cite{arbelaez2014multiscale,martin2001database} and CRF~\cite{krahenbuhl2013parameter} to produce pseudo masks, which are then adopted as ground-truth to train the segmentation network. However, such a pipeline often fails to generate accurate pseudo labels due to the gap between segmentation masks and object proposals. 
To overcome this limitation, graph-based learning was subsequently proposed to use the confident but a limited number of pixels mined from proposals as supervision. Compared to previous approaches, graph-based learning especially Graph Neural Network (GNN) can directly build long-distance edges between different nodes and aggregate information from multiple connected nodes, enabling to suppress the negative impact of the label noise. Besides, GNN performs well in semi-supervised tasks even with limited labels.

\begin{figure}
	\centering
	\includegraphics[width=\columnwidth]{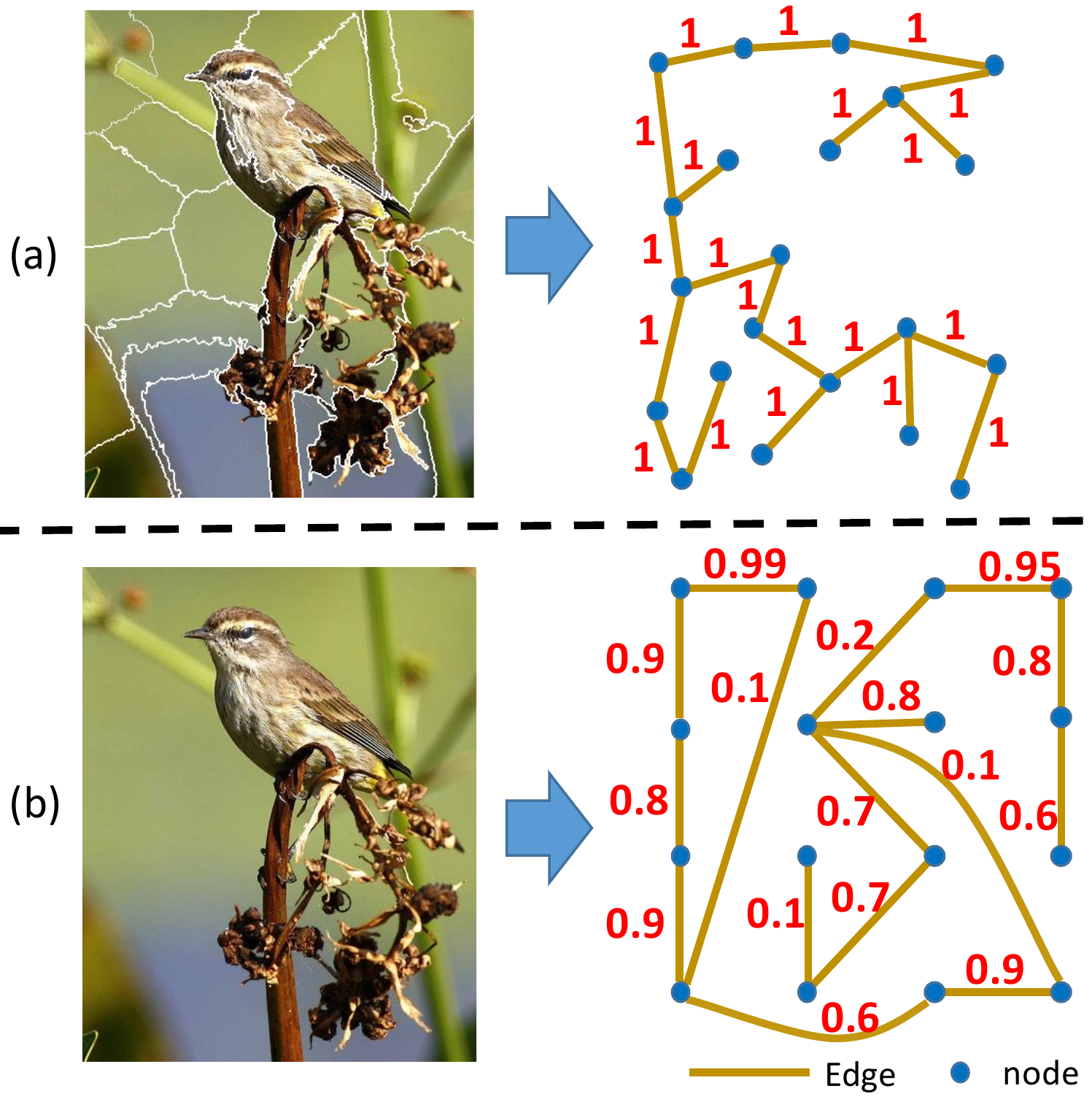}
	\caption{The difference between our built graph and that of previous approach~\cite{pu2018graphnet}. (a) Superpixel based approach~\cite{pu2018graphnet}. (b) Our approach. The numbers along the edges indicate the edge values, soft edge allows any edge weights between 0 and 1.}
	\label{img1}
\end{figure}

Recently, \emph{GraphNet}~\cite{pu2018graphnet} attempts to use Graph Convolutional Network (GCN)~\cite{kipf2016semi} for the BSSS task. They convert images to unweighted graphs by grouping pixels in a superpixel to a graph node~\cite{achanta2012slic}. Then the graph is input to a standard GCN with cross entropy loss to generate pseudo labels. However, there are two main drawbacks which limit its performance: (1) \emph{GraphNet}~\cite{pu2018graphnet} builds an unweighted graph as input, however, such a graph cannot accurately provide sufficient information since it treats all edges equally, with the edge weight being either 0 or 1, though in practice not all connected nodes expect the same affinity. (2) Using~\emph{GraphNet}~\cite{pu2018graphnet} will lead to incorrect feature aggregation as input nodes and edges are not 100\% accurate. For example, for an image that contains both dogs and cats, the initial node feature of dog fur and cat fur might be highly similar, which will produce some connected edges between them as edges are built based on feature similarity. Such edges will lead to a false positive case since \emph{GraphNet}~\cite{pu2018graphnet} only considers the initial edges for feature propagation. Thus, if the strong correlations among pixels from different semantics can be effectively alleviated, a better propagation model can be acquired to generate more accurate pseudo object masks. 

 

To this end, we design an Affinity Attention Graph Neural Network (\emph{$A^2$GNN}) to address the above mentioned issues. Specifically,  instead of using traditional method to build a unweighted graph, we propose a new affinity Convolutional Neural Network (CNN) to convert an image to a weighted graph. We consider that a weighted graph is more suitable than an unweighted one as it can provide different affinities for different node pairs. Fig.~\ref{img1} shows the difference between our built graph and that of the previous approach~\cite{pu2018graphnet}. It can be seen that the previous approach only considers locally connected nodes, and they build an unweighted graph based on superpixel~\cite{achanta2012slic}, while we consider both local and long distance edges, and the built weighted graph views one pixel as one node. 

Second, in order to produce accurate pseudo labels, we design a new GNN layer, in which both the attention mechanism and the edge weights are applied in order to ensure accurate propagation. 
So feature aggregation between pair-wise nodes with weak/no edge connection or low attention can be significantly declined, and thus eliminating incorrect propagation accordingly. The node attention dynamically changes as training goes on.

However, to guarantee the accuracy of supervision, we only choose a limited number of confident seed labels as supervision, which is insufficient for the network optimization. For example, only around 40\% foreground pixels are labeled in one image and none of them is 100\% reliable. To further tackle this issue, we introduce a multi-point (MP) loss to augment the training of \emph{$A^2$GNN}. Our MP loss adopts an online update mechanism to provide extra supervision from bounding box information. Moreover, in order to strengthen feature propagation of our \emph{$A^2$GNN}, MP loss attempts to close up the feature distance of the same semantic objects, making the pixels of the same object distinguishable from others. Finally, considering that the selected seed labels may not perfectly reliable, we introduce a consistency-checking mechanism to remove those noisy labels from the selected seed labels, by comparing them with the labels used in the MP loss. 

To validate the effectiveness of our \emph{$A^2$GNN}, we perform extensive experiments on PASCAL VOC. In particular, we achieve a new mIoU score of 76.5\% on the validation set. In addition, our \emph{$A^2$GNN} can be further smoothly transferred to conduct the bounding box supervised instance segmentation (BSIS) task or other weakly supervised semantic segmentation tasks. According to our experiments, we achieve new state-of-the-art or comparable performances among all these tasks.



Our main contributions are summarized as:

\begin{itemize}	
	\item We propose a new framework that effectively combines the advantage of CNN and GNN for weakly supervised semantic segmentation. To the best of our knowledge, this is the first framework that can be readily applied to all existing weakly supervised semantic segmentation settings and the bounding box supervised instance segmentation setting. 
	 
	\item We design a new affinity CNN network to convert a given image to an irregular graph, where the graph node features and the node edges are generated simultaneously. Compared to existing approaches, the graphs built from our method are more accurate for various weakly supervised semantic segmentation settings.
		
	\item We propose a new GNN, \emph{$A^2$GNN}, where we design a new GNN layer that can effectively mitigate inaccurate feature propagation through information aggregation based on edge weights and node attention. We further propose a new loss function (MP loss) to mine extra reliable labels using the bounding box constraint and remove existing label noise by consistency-checking.

	\item Our approach achieves state-of-the-art performance for BSSS on PASCAL VOC 2012 (\emph{val}: 76.5\%, \emph{test}: 75.2\%) as well as BSIS on PASCAL VOC 2012 ($\text{mAP}_{0.5}^r$: 59.1\%, $\text{mAP}_{0.7}^r$: 35.5\%, $\text{mAP}_{0.75}^r$: 27.4\%) and COCO ($\text{mAP}_{0.5}^r$: 43.9\%). 
	Meanwhile, when applying the proposed approach to other weakly supervised semantic segmentation settings, new state-of-the-art or comparable performances are achieved as well.
	
\end{itemize}

\section{Related Work}

\subsection{Weakly Supervised Semantic Segmentation}
According to the definition of supervision signals, weakly supervised semantic segmentation can be generally divided into the following categories: based on scribble label \cite{tang2018normalized,lin2016scribblesup,tang2018regularized}, bounding box label \cite{dai2015boxsup,khoreva2017simple,song2019box}, point label \cite{bearman2016s} and image-level class label \cite{ahn2018learning,wei2016stc,hou2018self}. Scribble, bounding box and point labels are stronger supervision signals compared to the image-level class label since both class and localization information are provided. Whereas image-level labels only provide image class tags with the lowest annotation cost. 

Different supervisions are processed with different methods to generate pseudo labels. For the scribble supervision, Lin \emph{et al.}~\cite{lin2016scribblesup} used superpixel based approach (\emph{e.g.}, SLIC \cite{achanta2012slic}) to expand the initial scribbles, and then used an FCN model \cite{long2015fully} to get the final predictions. Tang \emph{et al.} proposed two regularized losses~\cite{tang2018normalized,tang2018regularized} using the constraint energy loss function to expand the scribble information. For the point supervision, Bearman \emph{et al.}~\cite{bearman2016s} directly incorporated a generic object prior in the loss function. For image-level supervision, class activation map (CAM) \cite{zhou2016learning} is usually used as seeds to get pseudo labels. For example, Ahn and Kwak designed an affinity net~\cite{ahn2018learning} to obtain the transition probability matrix, and used random walk \cite{lovasz1993random} to get pseudo labels. Huang \emph{et al.}~\cite{huang2018weakly} proposed to use the seed region growing algorithm~\cite{adams1994seeded} to get pseudo labels from the initial confident class activation map. For the bounding box supervision, SDI~\cite{khoreva2017simple} used the segmentation proposal by combining MCG~\cite{pont2016multiscale} with GrabCut~\cite{rother2004grabcut} to generate the pseudo labels. Song \emph{et al.} proposed a box-driven method~\cite{song2019box}, using box-driven class-wise masking and filling rate guided adaptive loss to generate pseudo labels. Box2Seg~\cite{kulharia2020box2seg} attempt to design a segmentation network which is suitable to utilize the noisy labels as supervision.

Following the same definition, there are different level sub-tasks for weakly supervised instance segmentation: image-level~\cite{ahn2019weakly} and bounding box level~\cite{khoreva2017simple,hsu2019weakly}. For the image-level task, Ahn \emph{et al.} tried to generate the instance pseudo label using an affinity network ~\cite{ahn2019weakly}. For the bounding box task, SDI~\cite{khoreva2017simple} produced pseudo labels using the segmentation proposal by combining MCG \cite{pont2016multiscale} with GrabCut \cite{rother2004grabcut} while Hsu \emph{et al.}~\cite{hsu2019weakly} tried to design a new loss function which samples positive and negative pixels relying on bounding box supervision. 

\subsection{Graph Neural Network}

Generally, there are many different GNN \cite{abu2018n,guo2019attention,harada2018dual,kipf2016semi} methods designed for semi-supervised task and they have achieved  satisfying performances. Recently, GNN has been successfully used in various computer vision tasks such as person search~\cite{yan2019learning}, image recognition \cite{chen2019multi}, 3D pose estimation \cite{cai2019exploiting} and video object segmentation \cite{wang2019zero}, etc. 

For weakly supervised semantic segmentation, \emph{GraphNet} \cite{pu2018graphnet} was proposed for bounding box task using GCN \cite{kipf2016semi}. Specifically, based on the CAM technique~\cite{zhou2016learning}, bounding box supervision was converted to initial pixel-level supervision firstly. And then the superpixel method \cite{achanta2012slic} was applied to produce the graph nodes. They used a pre-trained CNN model to obtain the node feature, which was computed using average pooling for all its pixels. After that an adjacency matrix was computed based on the L1 distance between a node and its 8-neighbor nodes. Later, GCN was used to propagate the initial pixel-level labels to integral pseudo labels. Finally, all the pseudo labels were input into a semantic segmentation model for training. \emph{GraphNet} proved that GNN was one possible solution for weakly supervised semantic segmentation. However, this method has some limitations. First of all, using superpixels as nodes introduce incorrect node labels, and building an adjacency matrix by a threshold loses some important detailed information. Secondly, the performance of this method is limited by the usage of GNN \cite{kipf2016semi}, which only considers the non-weight adjacency matrix. Finally, only cross entropy loss is used for \emph{GraphNet}, which cannot mitigate the influence of incorrect nodes, edges and labels. 

In order to overcome the limitations of previous graph-based learning approach, we design a new approach, \emph{$A^2$GNN}, which takes a more accurate weighted graph as input and aggregates feature by considering both attention mechanism and edge weights. Meanwhile, we propose a new loss function to provide extra supervision and impose restrictions on the feature aggregation, thus our \emph{$A^2$GNN} can generate high quality pseudo labels.

\section{Generate Pixel-level Seed Label} \label{sec:CAM}
The common practice to initialize weakly supervised task is to generate pixel-level seed labels from weak supervision~\cite{pu2018graphnet, wang2020self,fan2020learning}. For the BSSS task, both image-level and bounding box-level labels are available. We use both of them to generate the pixel-level seed labels since image-level label can generate foreground seeds while bounding box-level label can provide accurate background seeds. To convert the image-level label to pixel-level labels, we use a CAM-based method~\cite{zhou2016learning, wang2020self, ahn2018learning, ahn2019weakly}. To generate pixel-level labels from bounding box supervision, Grab-cut~\cite{rother2004grabcut} is used to generate the initial labels, and the pixels which do not belong to any box are regarded as background labels. Finally, these two types of labels are fused together to generate the pixel-level seed labels.

Specifically, we use SEAM~\cite{wang2020self}, which is a self-supervised classification network, to generate the pixel-level seed labels from image-level supervision. Suppose a dataset with category set $C = [c_0,c_1,c_2,...,c_{N-1}]$, in which $c_0$ is background with the rest representing foreground categories. The pixel-level seed labels from image-level supervision are:
\begin{equation}
	M_I = \text{Net}_{\text{SEAM}}(I),\label{eq:M_I}
\end{equation}
where $M_I$ is the generated seed labels. $\text{Net}_{\text{SEAM}}(\cdot)$ is the classification CNN used in SEAM~\cite{wang2020self}.
 
For the BSSS task, as it provides bounding box-level label in addition to image-level label. We also generate pixel labels from the bounding box label as it can provide accurate background labels and object localization information. Given an image, suppose the bounding box set is $B = \left \{ B_1,...,B_M\right \}$. For a bounding box $B_k$ with label $L_{B_k}$, its height and width are $h$ and $w$, respectively. We use Grab-cut~\cite{rother2004grabcut} to generate the seed labels from bounding box supervision, the seed labels for each bounding box are defined as:
\begin{equation}
M_{B_k}(i) = \begin{cases}
\text{Grab}(i),  & \text{if }  i \in B_k  \text{ and } $$ \text{Grab}(i) \neq c_0 \\
255, & \text{else}
\end{cases}, \label{eq:B_k}
\end{equation}
where $\text{Grab}(\cdot)$ is the Grab-cut operator and 255 means the pixel label is unknown.

Pixels not belonging to any bounding box are expressed as background, and the final seed labels generated from bounding box are:
\begin{equation}
M_{B}(i) = \begin{cases}
c_0, & \text{if }  i \notin B  \\
M_{B_k}(i),  & \text{if }  i \in B
\end{cases}. \label{MB}
\end{equation}

\begin{figure}
	\centering
	\includegraphics[width=\columnwidth]{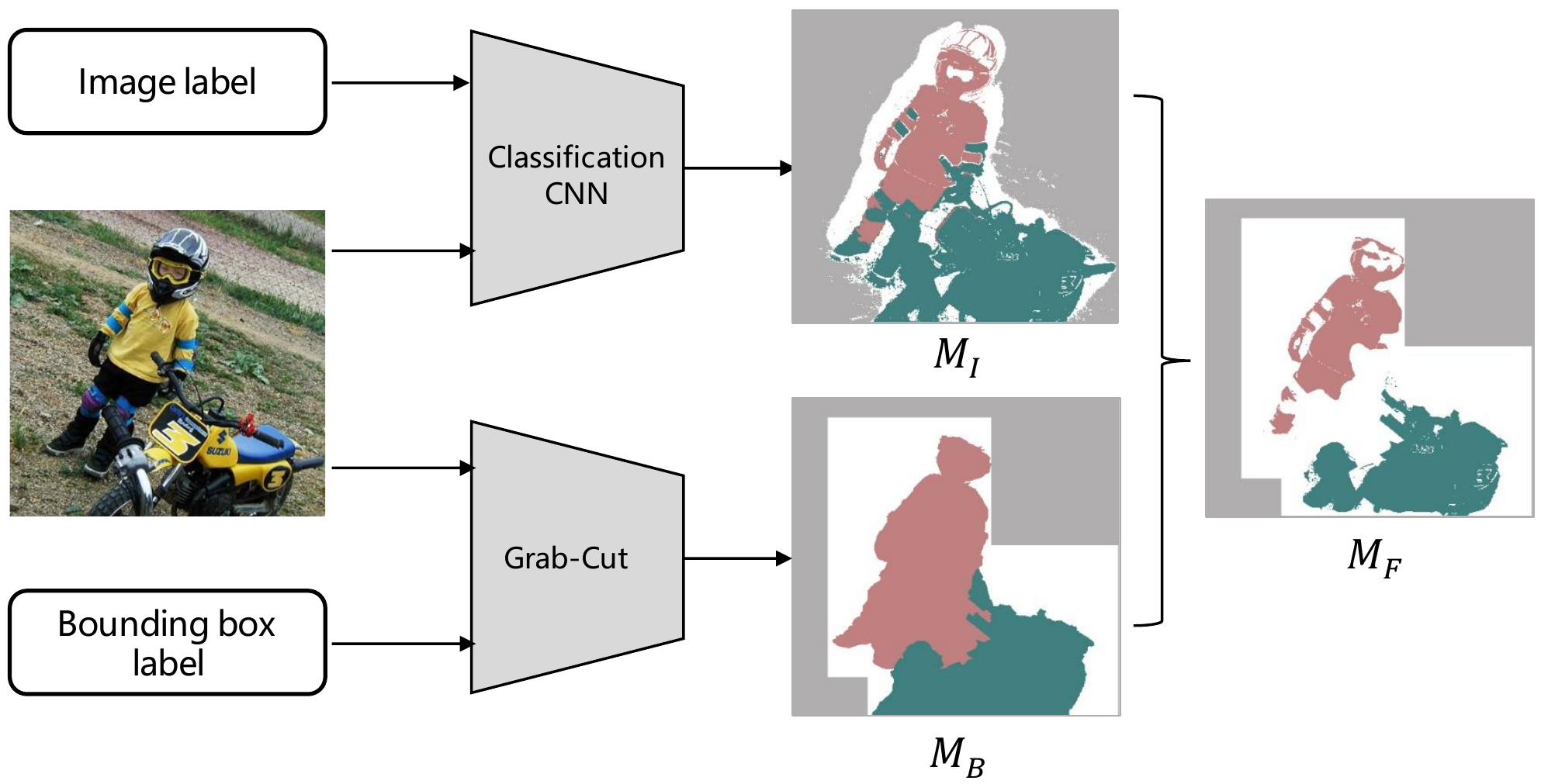}
	\caption{An example of generating pixel-level seed labels. Given an image with its label, we firstly generate $M_{I}$ from image-level label using a classification CNN and SEAM~\cite{wang2020self} method. Meanwhile, bounding box label is transferred to pixel-level label $M_{B}$ using Grab-cut. Finally, $M_{I}$ and $M_{B}$ are integrated together to get the pixel-level seed label $M_{F}$. Each color represents one class and ``white" means the pixel label is unknown.}
	\label{img2}
\end{figure}

For pixel $i$ in the image, the final pixel-level seed label is defined as :
\begin{equation}
\resizebox{0.9\columnwidth}{!}{$
	M_{F}(i)=\begin{cases}
	M_B(i), & \text{if }  i \notin B\\ 
	M_{I}(i), & \text{if } i \in B \text{ and } M_{B}(i) = M_{I}(i)\\
	M_{B_k}(i), & \text{if } i \in B_k \text{ and } L_{B_k} \notin S(M_{I\text{-}B_k}) \\ 
	255, & \text{else} 
	\end{cases},\label{MF}
$}
\end{equation}
where $S(M_{I\text{-}B_k})$ is the set of predicted categories in ${M_I}$ for bounding box $B_k$. $L_{B_k} \notin S(M_{I\text{-}B_k})$ indicates that there is no correct predicted label in $M_I$ for bounding box $B_k$ and we therefore use the prediction from $M_{B_k}$ as the final seed labels.

In Fig.~\ref{img2}, an example is given to demonstrate the process to convert bounding box supervision to pixel-level seed labels. After combing $M_{I}$ and $M_{B}$, we can get the pixel-level seed label.

\section{The Proposed $A^2$GNN}\label{sec:GNN}
\subsection{Overview}
\begin{figure*}
	\centering
	\includegraphics[width=\textwidth]{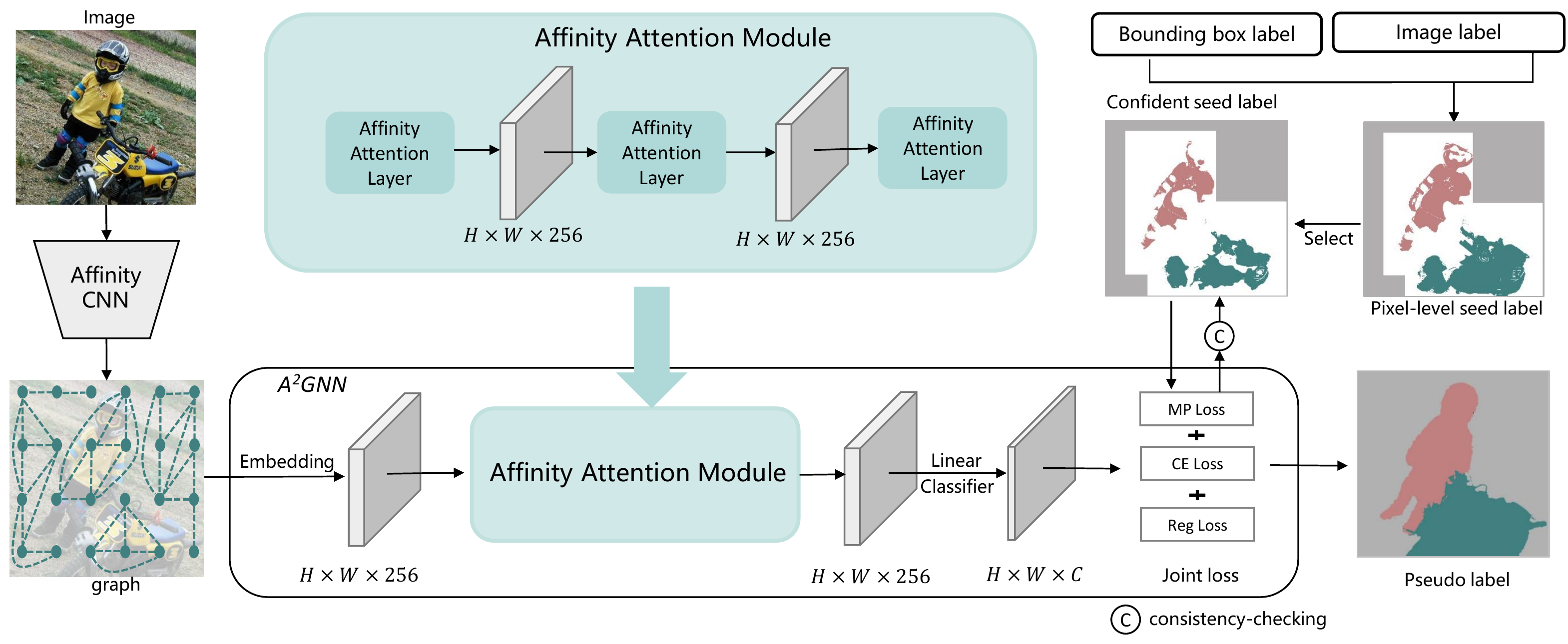}
	\caption{The framework of our proposed \emph{$A^2$GNN}. Firstly, we generate pixel-level seed labels using the bounding box and the image-level label. Then our affinity CNN is used to convert images to graphs. Meanwhile, we select confident labels from pixel-level seed labels as the node labels (The node labels in the white region are unknown). Finally, \emph{$A^2$GNN} uses the graph data as input and the node labels as supervision to produce pseudo labels.}
	\label{fig:wholeframe}
\end{figure*}
In order to utilize GNN to generate the accurate pixel-level pseudo labels, there are three main problems: (1) How to provide useful supervision information and reduce the label noise as much as possible. (2) How to convert the image data to accurate graph data.  (3) How to generate accurate pseudo labels based on the built graph and the supervision.
  
In this section, we will elaborate on the proposed \emph{$A^2$GNN} to address the above mentioned three main problems. To generate an accurate graph, we propose a new affinity CNN to convert an image to a graph. To provide accurately labeled nodes for the graph, we select highly confident pixel-level seed labels as node labels, and at the same time, we introduce extra online updated labels based on the bounding box supervision, meanwhile, the pixel-level seed labels are further refined by consistency-checking. To generate accurate pseudo labels, we design a new GNN layer since the previous GNN, such as GCN~\cite{kipf2016semi} or AGNN~\cite{guo2019attention} is designed based on the assumption that labels are 100\% accurate, while in this case, there is no foreground pixel label being 100\% reliable.  

In Fig.~\ref{fig:wholeframe}, we show the main process of our approach, which can be divided into three steps:

\begin{enumerate}[(1)]
	\item Generating confident seed labels. In this step, both image-level labels and bounding box-level labels are converted to initial pixel-level seed labels, as explained in section~\ref{sec:CAM}. Then the pixel-level seed labels with high confidence will be selected as confident seed labels (section~\ref{sec:sup}).
	
	\item Converting images to graphs. In this step, we propose a new affinity CNN to generate the graph. Meanwhile, the selected confident seed labels will be converted to corresponding node labels.
	
	\item Generating final pixel-level pseudo labels. \emph{$A^2$GNN} is trained using the converted graph as input, and it makes the prediction for all nodes in the graph. After converting node pseudo labels to pixel labels, we generate the final pixel-level pseudo labels. 		
\end{enumerate}
 
After that, a FCN model such as Deeplab~\cite{chen2017deeplab,chen2017rethinking} for BSSS or MaskR-CNN~\cite{he2017mask} for BSIS is trained using above pixel-level pseudo labels as supervision.

In the following section, we will first introduce how to provide useful supervision, and then we will give an explanation about how to build a graph from the image (section~\ref{sec:graph}). Finally, we will introduce \emph{$A^2$GNN}, including its affinity attention layer (section~\ref{sec:AAL}) and its loss function (section~\ref{loss}).

\subsection{Confident Seed Label Selection}\label{sec:sup}

An intuitive solution is to use the pixel-level seed label $M_F$ obtained from Eq.~(\ref{MF}) as the seed labels. However, $M_F$ is noisy and directly using it will be harmful to train a CNN/GNN. As a result, in this paper, we only select those highly confident pixel-level seed labels in $M_{F}$ as the final seed labels. 
Specifically, we use a dynamic threshold to select top 40\% confident pixel labels $M'_{I}$ following~\cite{zhang2019reliability} from the pixel label $M_I$ in Eq.~(\ref{eq:M_I}). Then the selected seed labels are defined as:
\begin{equation}
\resizebox{0.87\columnwidth}{!}{$
	M_g(i)= \begin{cases}
	M_F(i), & \text{if } M_F(i) = M_B(i) \text{ or } M_F(i) = M'_I(i)\\
	255,& \text{else} 
	\end{cases}, \label{LVl}
	$}
\end{equation}  
where 255 means that the label is unknown. $M_B$ and $M_F$ are obtained from Eq.~(\ref{MB}) and Eq.~(\ref{MF}), respectively. Fig.~\ref{fig:wholeframe} (top-right) illustrates the confident label selection.

Although noisy labels can be removed considerably, the confident label selection has two main limitations: 1) it also removes some correct labels, making the rest labels scarce and mainly focus on discriminative object parts (\eg, human head) rather than uniformly distributed in the object; 2) there still exist non-accurate labels. 

To tackle the label scarcity problem in the BSSS task, we propose to mine extra supervision information from the available bounding box. Assuming all bounding boxes are tight, for a random row or column pixels inside a bounding box, there is at least one pixel belonging to the object. Identifying these nodes can provide extra foreground labels. And using the online updated labels, we introduce a new consistency-checking mechanism to further remove some noisy labels from $M_g$. We will describe the detailed process in section~\ref{loss} since they rely on the output of our \emph{$A^2$GNN}.

\subsection{Graph Construction}\label{sec:graph}
\subsubsection{Affinity CNN}
We propose a new affinity CNN to produce an accurate graph from an image using the available affinity labels as supervision. This is because affinity CNN has the following merits. First, instead of regrading one superpixel as a node, it views a pixel as one node which introduces less noise. Second, the affinity CNN uses node affinity labels as training supervision, which ensures to generate suitable node features for this specific task, while previous \emph{GraphNet} \cite{pu2018graphnet} uses classification supervision for training. Third, compared to the short distance unweighted graph (edges are only represented as 0 and 1) built in \emph{GraphNet}~\cite{pu2018graphnet}, an affinity CNN can build a weighted graph with soft edges covering a long distance, which gives more accurate node relationship.  
 
Different from prior works~\cite{ahn2018learning,ahn2019weakly,song2020cian,wu2019improving} that use all noisy labels in Eq.~(\ref{MF}) as supervision, our affinity CNN only uses the confident seed labels as defined in Eq.~(\ref{LVl}) as supervision to predict the relationship of different pixels. 

In order to train our affinity CNN, we firstly generate class-agnostic labels from the confident pixel-level seed labels $M_g$ from Eq.~(\ref{LVl}):
\begin{equation}
\resizebox{0.87\columnwidth}{!}{$
	L_A(i,j)=\left\{\begin{matrix}
	1, &  (i,j) \in R_{pair} \text{ and } M'_{g}(i) = M'_{g}(j) \\  
	0, &  (i,j) \in R_{pair} \text{ and } M'_{g}(i) \neq M'_{g}(j) \\ 
	255, & else
	\end{matrix}\right.
	$}
\label{aff},
\end{equation}
where both $i$ and $j$ are pixel indices and $255$ means that this pixel pair is not considered. $M'_{g}$ is the down-sampled result of $M_g$ in order to keep the same height and width with the feature map. $R_{pair}$ is the pixel pair set to train the affinity CNN, and it satisfies the following formula:
\begin{equation}
\begin{aligned}
R_{pair}=\left\{(i,j)|M'_{g}(i) \neq 255 \text{ and } M'_{g}(j) \neq 255  
\right.  \\
\text{and }  \left ||Pos(i)-Pos(j)||_2 \leqslant  r \right\} , \label{Raff}
\end{aligned}
\end{equation}
where $||\cdot||_2$ is an Euclidean distance operator, $Pos(\cdot)$ represents the coordinate of the pixel. $r$ is the radius, which is used to restrict the selection of a pixel pair. 

Given an image $I$, suppose the feature map from the affinity CNN is $F_{\text{A}}$, following \cite{ahn2018learning}, L1 distance is applied to compute the relationship of the two pixels $i$ and $j$ in $F_{\text{A}}$:
\begin{equation}
D(i,j)=exp(-\frac{\left \| F_A(i)-F_A(j) \right \|}{d_A}),\label{Eq:D}
\end{equation}
where $d_A$ is the channel dimension of feature map $F_A$. 

The training loss of affinity CNN is defined as:
\begin{equation}
	\mathcal{L}_{\text{\tiny{Aff}}} = \mathcal{L}_{\text{\tiny{Ac}}} + \lambda \mathcal{L}_{\text{\tiny{Ar}}}. \label{L_A}
\end{equation}

In Eq.~(\ref{L_A}), $\mathcal{L}_{\text{\tiny{Ac}}}$ is a cross-entropy loss which focuses on using the annotated affinity labels as supervision:
\begin{equation}
\resizebox{0.87\columnwidth}{!}{$
	\begin{aligned}
	\mathcal{L}_{\text{\tiny{Ac}}}=&-\frac{1}{|A^+|}\sum\limits_{(i,j) \in A^+ }L_A(i,j)log(D(i,j))\\ 
	&-\frac{1}{|A^-|}\sum\limits_{(i,j) \in A^-}(1-L_A(i,j))log(1-D(i,j)),\label{LGA} 
	\end{aligned}
	$}
\end{equation}
where $A^+$ is the node pair set with $L_A(i,j)=1$, $A^-$ is the node pair set with $L_A(i,j)=0$. Operator $|\cdot|$ defines the number of elements. 

Note that only using the confident labels as supervision is insufficient to train a CNN when only considering $\mathcal{L}_{\text{\tiny{Ac}}}$ as loss function. In order to expand the labeled region to unlabeled region, we propose an affinity regularized loss $\mathcal{L}_{\text{\tiny{Ar}}}$ to encourage propagating from labeled pixels to its connected unlabeled pixels. In other words, instead of only considering pixel pairs in $R_{pair}$, we consider all pixel pairs which satisfy the following formula:
\begin{equation}
	R_{Ar} =\left\{(i,j)| \left \| Pos(i)-Pos(j) \right \|_2 \leqslant  r \right\}.\label{Rad}
\end{equation}

Then the affinity regularized loss is defined as:
\begin{equation}
\mathcal{L}_{Ar} = \sum_{i=1}^{\text{HW}}\sum_{(i,j) \in R_{Ar}}G(i,j) \frac{\left \| (F_A(i) - F_A(j)) \right \|}{d_A}, \label{Gd}
\end{equation}
where $G(\cdot,\cdot)$ is a Gaussian bandwidth filter \cite{tang2018normalized}, which utilizes the color and spatial information:
\begin{equation}
\resizebox{0.87\columnwidth}{!}{$
	G(i,j) = exp(-\frac{\left \| Pos(i)-Pos(j)\right \|_2}{2\sigma_{xy}^2 }-\frac{\left \| Cor(i)-Cor(j)\right \|_2}{2\sigma_{rgb}^2 }) \cdot \left  [ i \neq j\right ],\label{Kab}
	$}
\end{equation}
where $Pos(i)$ and $Pos(j)$ are the spatial positions of $i$ and $j$, respectively. $Cor(\cdot)$ is the color information and $[\cdot]$ is Iverson bracket.

\begin{figure}
	\centering
	\includegraphics[width=\columnwidth]{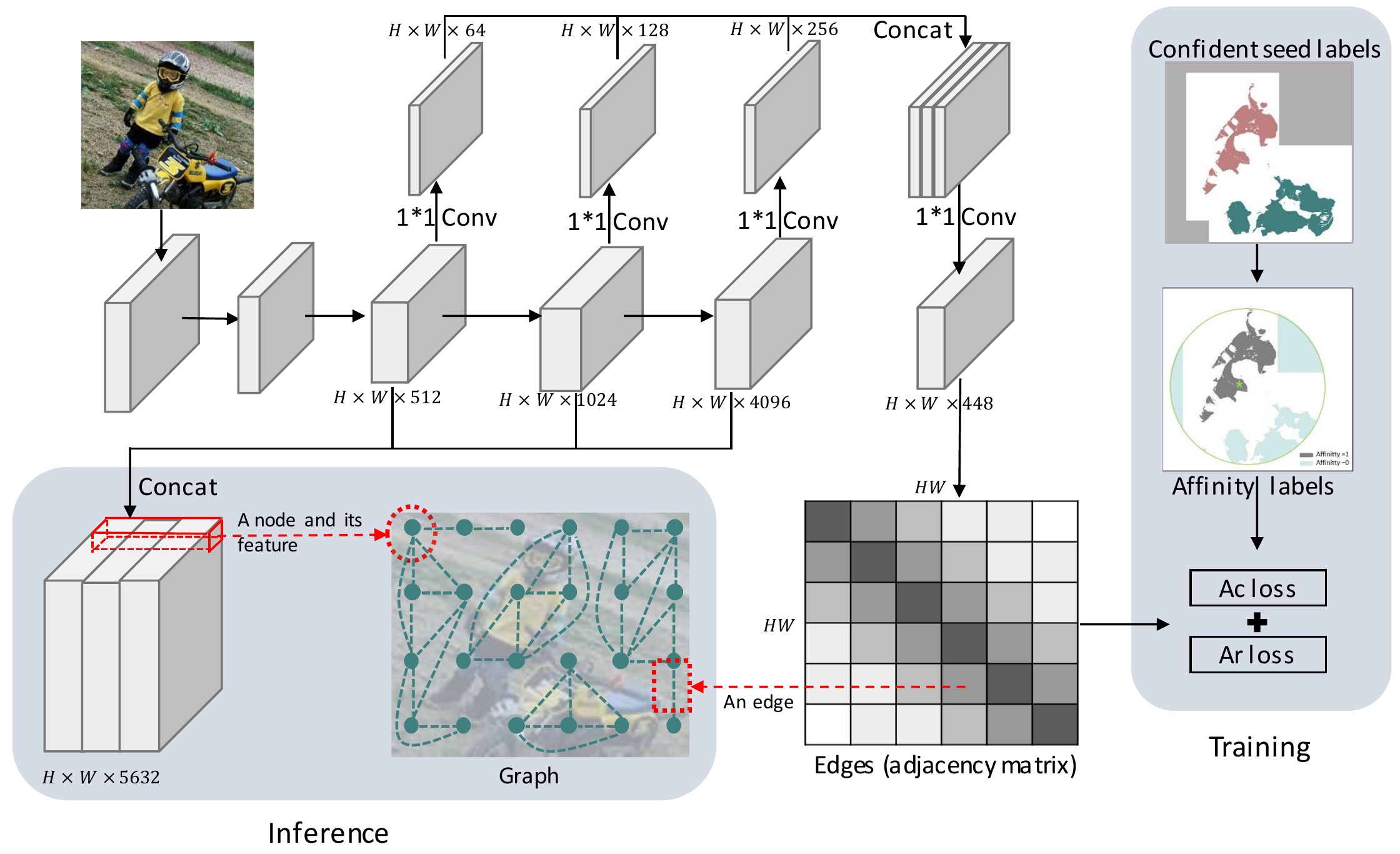}
	\caption{Converting an image to a graph using our affinity CNN. During inference, the given image will be converted to a graph, in which a node is a pixel in the concatenated feature maps from the last three blocks and its feature is the corresponding pixel feature. The weight of graph edges is defined as the predicted affinity and they are represented as an adjacency matrix, in which each row corresponds to all edges between one node and all nodes.}
	\label{fig:AffinityNet}
\end{figure}

\subsubsection{Convert Image to Graph}\label{sec:image2graph}
Usually, a graph is represented as $G = (V, E)$ where $V$ is the set of nodes, and $E$ is the set of edges. Let $v_i \in V$ denote a node and $E_{i,j}$ represents the edge between $v_i$ and $v_j$. $X \in \mathbb{R}^{N_g*D_g}$ is a matrix representing all node features, where $N_g$ is the number of nodes and $D_g$ is the dimension of the feature. In $X$, the \emph{i}th feature, represented as $x_i$, corresponds to the feature of node $v_i$. The set of all labeled nodes is defined as $V^l$, and the set of remaining nodes is represented as $V^u$, and $V= V^l \cup  V^u$.

During training, our affinity CNN uses the class-agnostic affinity labels as supervision and learns to predict the relationship of pixels. During inference, given an image, our affinity CNN will output $V$, $X$ and $E$ simultaneously for a graph as shown in Fig.~\ref{fig:AffinityNet}. Specifically, the node $v_i$ and its feature $x_i$ corresponds to \emph{i}th pixel and its all-channel features in the concatenated feature map from the backbone. For two nodes $v_i$ and $v_j$, their edge $E_{ij}$ is defined as:
\begin{equation}
E_{ij} = \begin{cases}
D(i,j), & \text{if } D(i,j)>\sigma\\ 
0,& \text{else} 
\end{cases}, \label{Eab}
\end{equation} 
where $i$ and $j$ are pixels in the feature map, and $D(i,j)$ is obtained from Eq.~(\ref{Eq:D}). Here we use a threshold $\sigma$ (set as 1$e$-3 in our experiment) to make some low affinity edges be $0$. Finally, we generate the normalized features:
\begin{equation}
x_{i,j} = x_{i,j} / \sum_{j=1}^{D_g}(x_{i,j}), \label{nor}
\end{equation} 
where  $x_{i,j}$ represents the $j$th value of feature $x_i$ and $D_g$ is the feature dimension.

\subsection{Affinity Attention Layer}\label{sec:AAL}

Effective GNN architectures have been studied in existing works ~\cite{kipf2016semi,guo2019attention}, where most of them are designed based on the assumption that the graph node and edge information is 100\% accurate. However, in the BSSS task, it is not the case. We propose a new GNN layer with attention mechanism to mitigate this issue. As shown in Fig.~\ref{fig:wholeframe}, in the proposed \emph{$A^2$GNN}, an affinity attention module is applied after the embedding layer. The affinity attention module includes three new GNN layers named affinity attention layers. Finally, an output layer is followed to predict class labels for all nodes.

Specifically, we use a feature embedding layer followed by a ReLU activation function in the first layer to map the initial node features to the same dimension of the assigned feature:
\begin{equation}
H^{1}=ReLU(XW^0),
\end{equation}
where $X$ is the feature matrix defined in section~\ref{sec:image2graph} and $W^0$ is the parameter set of the embedding layer. Then we design several affinity attention layers to leverage the edge weights:
\begin{equation}
H^{l+1}=P^lH^l,
\label{aff-atten}
\end{equation}
where $P^l \in \mathbb{R}^{N_G\times N_G}$, $N_G$ is the number of nodes. For node $v_i$, the affinity attention $P^l(i,j)$ from node $v_j$ is defined as:
\begin{equation}
\resizebox{0.87\columnwidth}{!}{$
\begin{aligned}
P^l(i,j)& = \text{softmax}(w^l \cos(H^l(i), H^l(j))+\beta [\cos(H^l(i), H^l(j))>0] E_{ij})\\
&=\frac{\exp\left \{w^l \cos(H^l(i), H^l(j))+\beta [\cos(H^l(i), H^l(j))>0] E_{ij}\right \}}{\sum\limits_{v_j\in S(i)}\exp\left \{w^l\cos(H^l(i), H^l(j))+\beta [\cos(H^l(i), H^l(j))>0] E_{ij}\right \}}, \label{Pl}
\end{aligned}
$}
\end{equation}
where $l \in \left\{1,2,..., L\right\}$ is the layer index ($L$ is set as $3$ in our model) of \emph{$A^2$GNN} and $w^l$ is the learning parameter. $S(i)$ is the set of all the nodes connected with $v_i$ (including itself). $\left [ \cdot \right ]$ equals 1 when $\cos(\cdot,\cdot)>0$ and otherwise equals 0. $H^l(i)$ and $H^l(j)$ correspond to the features of $v_i$ and $v_j$ at layer $l$, respectively. $\cos(\cdot, \cdot)$ is used to compute the cosine similarity, which is a self-attention module. $E_{ij}$ is the predicted edge in Eq.~(\ref{Eab}). $\beta$ is a weighting factor. The final output  is:
\begin{equation}
O = \text{softmax}(H^{L+1}W^{L+1})\label{AO},
\end{equation}
where $W^{L+1}$ is the parameter set of the output layer. 

\begin{figure}
	\centering
	\includegraphics[width=0.8\columnwidth]{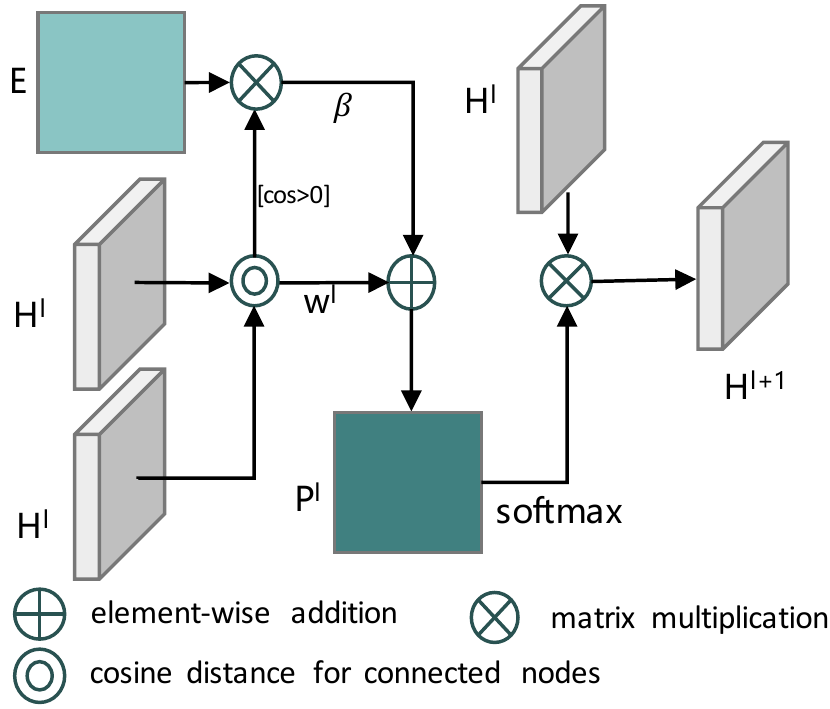}
	\caption{Our proposed affinity attention layer. $E$ is the adjacent matrix which provides soft edges information. $H^l$ is the input feature of the layer and $H^{l+1}$ is the output feature. $P^l$ is the computed affinity attention matrix. $w^l$ is the learning parameter and $\beta$ is the weighting factor. With the attention mechanism and the soft edges, it can ensure accurate feature propagation.}
	\label{fig:GNNlayer}
\end{figure}

Fig.~\ref{fig:GNNlayer} shows the flowchart of our affinity attention layer. Compared to GCN layer~\cite{kipf2016semi} and AGNN layer~\cite{thekumparampil2018attention}, our affinity attention layer makes full advantage of node similarity and edge weighting information.

\subsection{Training of $A^2$GNN}\label{loss}

As described in section~\ref{sec:sup}, we only select confident labels as supervision, which is insufficient for the network optimization. In order to address this problem, we impose multiple supervision on our \emph{$A^2$GNN}. Specifically, we design a new joint loss function, including a cross-entropy loss, a regularized shallow loss \cite{tang2018regularized} and a multi-point (MP) loss:
\begin{equation}
\mathcal{L}_{G} = \mathcal{L}_{ce} + \mathcal{L}_{mp} + \lambda_1 \mathcal{L}_{reg}, \label{eq:L_G}
\end{equation}
where $\mathcal{L}_{ce}$ is the cross-entropy loss to use the labeled nodes $M_g$ generated in section~\ref{sec:sup}. $\mathcal{L}_{reg}$ is the regularized loss using the shallow feature, \ie, color and spatial position. $\mathcal{L}_{mp}$ is the newly proposed MP loss to use the bounding box supervision.

\subsubsection{Cross Entropy Loss}
$L_{ce}$ is the cross entropy loss, which is used to optimize our \emph{$A^2$GNN} based on the labeled nodes $M_g$: 
\begin{equation}
\mathcal{L}_{ce} = -\frac{1}{|V^l| }\sum_{\begin{subarray}{l}
	c_{i}\in C \\v_j \in V^l
	\end{subarray}} [c_i = M_g(v_j)]log(O^{c_i}(v_j)),\label{GLCE}
\end{equation}
where $V^l$ is the set of all labeled nodes. $M_g(v_j)$ means the label of node $v_j$. $\left |\cdot \right|$ is used to compute the number of elements. $O^{c_i}(v_j)$ is the predicted probability of being class $c_i$ for node $v_j$. $V^l$ is the set of labeled nodes.

\subsubsection{Regularized Loss}
$\mathcal{L}_{reg}$ is the regularized loss which explores the shallow features of images. Here we use it to pose constraints based on the image-domain information (\eg, color and spatial position).   
\begin{equation}
\mathcal{L}_{reg} = \sum_{\begin{subarray}{l}
	c_{i}\in C \\v_a \in V
	\end{subarray} }\sum_{\begin{subarray}{l}
	c_{j}\in C \\v_b \in V
	\end{subarray} } G(v_a,v_b)O^{c_i}(v_a)O^{c_j}(v_b). \label{GReg}
\end{equation}

In Eq.~(\ref{GReg}), $v_a$ and $v_b$ represent two graph nodes. $V$ is the set including all nodes, and $G(v_a,v_b)$ is defined in Eq.~(\ref{Kab}).

\subsubsection{Multi-Point Loss}\label{sec:MP}
Inspired by \cite{hsu2019weakly}, we design a new loss term named multi-point (MP) loss to acquire extra supervision from bounding boxes. This is because the labeled nodes generated in section~\ref{sec:sup} are scarce and not perfectly reliable, which could be complemented by the bounding box information. The MP loss is based on the following consideration. Assuming all bounding boxes are tight, for a random row or column pixels inside one bounding box, there is at least one pixel belonging to the object, if we can find out all these nodes, then we can label them with the object class and close up their distance in the embedding space. Thus, MP loss makes the object easy to be distinguished. 

Specifically, for each row/column in the bounding box, the node with the highest probability to be classified to the bounding box class label is regarded as the selected node. Following the same definition in section~\ref{sec:CAM}, suppose the bounding box set in one image is $B$, then for arbitrary bounding box $B_j$ in $B$, firstly we need to select the highest probability pixel for each row/column:

\begin{equation}
i_{\text{max}}^{l_m} = \text{index}(\max_{i \in B_{j}^{l_m}}(O^{B_j}(i))), \label{eq:index}
\end{equation}
where $l_m$ means the \emph{m}th row/column,  $\max\limits_{i \in B_{j}^{l_m}}(O^{B_j}(i))$ means that for each row/column, we select the node which has the highest probability to be classified as the same label with $B_j$, $\text{index}\left ( \cdot \right )$ returns the index of selected node. $i_{\text{max}}^{l_m}$ is the index of the selected node in the \emph{m}th row/column.

Then the set that contains all selected nodes for the bounding box $B_j$ are defined as $K_j$:
\begin{equation}
K_j=\left \{ i_{\text{max}}^{l_1}, i_{\text{max}}^{l_2},i_{\text{max}}^{l_3},...,i_{\text{max}}^{l_{(w+h)}} \right \},\label{eq:K}
\end{equation}
where $w$ and $h$ represents the width and height of $B_j$, respectively. Then all selected nodes for all bounding boxes are defined as $K$:
\begin{equation}
K = \left \{ K_1, K_2,...,K_M \right \},
\end{equation}
where $M$ is the number of bounding boxes. Finally, the MP loss is defined as:

\begin{equation}
\begin{aligned}
\mathcal{L}_{mp} &= -\frac{1}{N_p}\sum_{K_j \in K}\sum_{k_i \in K_j}log(O^{B_j}(k_i)) +\\
&\frac{1}{N_f}\sum_{K_j \in K}\sum_{\begin{subarray}{l}
	k_m \in K_j \\k_n \in K_j
	\end{subarray}}[k_m \neq k_n](d(H(k_m), H(k_n))).\label{Lmp}
\end{aligned}
\end{equation} 

In Eq.~(\ref{Lmp}), $d(\cdot, \cdot)$ is used to compute feature distance, where we set $d(\cdot, \cdot) = 1 - \cos(\cdot, \cdot)$. $H(k_m)$ and $H(k_n)$ correspond to the features from the last affinity attention layer $H^{L+1}$ for node $k_m$ and $k_n$, respectively. Both $N_p$ and $N_f$ are the number of sum items. MP loss tries to pull the selected nodes closer in the embedding space, while all other nodes connecting with them will benefit from this loss. This is because GNN layer can be regarded as a layer to aggregate features from the connected nodes, it will encourage the other connected nodes to share a similar feature with them. In other words, MP loss will make the nodes belonging to the same object easy to be distinguished since they are assigned to a similar feature in the embedding space. In our model, we only enforce MP loss on $K_j$ (Eq.~(\ref{eq:K})) rather than all labeled nodes. This is because other nodes from $M_g$ still have noisy labels, and at the same time, the labeled foreground nodes in $M_g$ focus more on discriminative parts of the object.


\subsubsection{Consistency-Checking}\label{sec:cc}
As mentioned in section~\ref{sec:sup}, although we select some confident seed labels as supervision, noisy labels are still inevitable. Considering that we provide some extra online labels in our MP loss, which selects the highest probability pixel in each row/column in the box as additional labels, we assume that most additional labels in MP loss are correct, then for each box, we first generate a prototype using the feature of all additional labels inside the box:
\begin{equation}
	H_{P}^{K_j} = \frac{1}{N_{K_j}} \sum_{k_i \in K_j}H(k_i), \label{eq:F_P}
\end{equation}
where $H_{P}^{K_j}$ represents the prototype of the $j$th bounding box and $N_{K_j}$ is the number of the selected pixel.

Then for each bounding box, we compute the distance between all selected confident seed labels in Eq.~(\ref{LVl}) and the prototype, and finally the seed labels which are far away from the prototype are considered as noisy label and removed in each iteration:
\begin{equation}
	M_g^{K_j}(i) = \left\{\begin{matrix}
	M_g^{K_j}(i), & \text{if } d(H^{K_j}(i), H_{P}^{K_j}) >0 \\ & \text{ and } M_g^{K_j}(i) = L_{B_j}\\ 
	
	255, & \text{else}
	\end{matrix}\right.,
\end{equation}
where $M_g^{K_j}$ is the selected confident label map of the $j$th bounding box (section~\ref{sec:sup}), $H^{K_j}$ is the corresponding feature map for the $j$th bounding box from the last affinity attention layer $H^{L+1}$, and $d(\cdot, \cdot)$ is the operator to compute the cosine distance.

\begin{table*}[]
	\caption{Comparison with other approaches on PASCAL VOC 2012 \emph{val} and \emph{test} sets for BSSS. F: fully supervised. S: scribble supervised. B: bounding box supervised. Seg.: fully-supervised segmentation model}
	\centering
	\begin{tabular}{lccccccc}
		\hline
		\multirow{2}{*}{Method} & \multirow{2}{*}{Pub.} & \multirow{2}{*}{Seg.} & \multirow{2}{*}{Sup.} & \multicolumn{2}{c}{val mIoU (\%)}  & \multicolumn{2}{c}{test mIoU (\%)} \\ \cline{5-8} 
		& &   &  & \multicolumn{1}{c}{w/o CRF} & \multicolumn{1}{c}{w/ CRF} & \multicolumn{1}{c}{w/o CRF} & \multicolumn{1}{c}{w/ CRF} \\ \hline
		Deeplab-V1 \cite{chen2014semantic} & -& - & F & 62.3& 67.6 &-&70.3\\	
		Deeplab-Vgg~\cite{chen2017deeplab} & TPAMI'18& -& F &68.8&71.5&-&72.6\\
		Deeplab-Resnet101~\cite{chen2017deeplab} & TPAMI'18& -&F&75.6&76.8&-&79.7\\
		PSPNet~\cite{zhao2017pspnet}& CVPR'17&- & F& 79.2& - &82.6&-\\
		ScribbleSup~\cite{lin2016scribblesup}&CVPR'16 & Deeplab-Vgg& S& -& 63.1&-  &- \\
		RAWKS~\cite{vernaza2017learning}&  CVPR'17 &Deeplab-V1&S&-&61.4&-&-\\
		Regularized Loss~\cite{tang2018regularized}&ECCV'18 &Deeplab-Resnet101& S     & 73.0&75.0&-&-\\ \hline
		Box-Sup~\cite{dai2015boxsup} & CVPR'15 &Deeplab-V1&B&-& 62.0& -  & 64.6\\
		WSSL~\cite{papandreou1502weakly}& CVPR'15 &Deeplab-V1& B & -& 60.6 &-  &62.2\\
		GraphNet~\cite{pu2018graphnet}& ACMM'18 &Deeplab-Resnet101& B & 61.3   & 65.6 &-  &-\\
		SDI~\cite{khoreva2017simple}& CVPR'17  &Deeplab-Resnet101& B & -  & 69.4 &-  &-\\
		BCM~\cite{song2019box}& CVPR'19&Deeplab-Resnet101&B&-&70.2&-&-\\
		Lin \emph{et al.}~\cite{li2018weakly}& ECCV'18 &PSPNet& B &-& 74.3 &-&-\\
		Box2Seg~\cite{kulharia2020box2seg}& ECCV'20 &UperNet~\cite{xiao2018unified}& B &74.9&76.4&-&-\\
		Box2Seg-CEloss~\cite{kulharia2020box2seg}& ECCV'20 &UperNet~\cite{xiao2018unified}& B&72.7&- &-&-\\ \hline
		\emph{$A^2$GNN} (ours)& -  &Deeplab-Resnet101& B & \textbf{72.2}   & \textbf{73.8} & \textbf{72.8}  & \textbf{74.4} \\
		\emph{$A^2$GNN} (ours)& -  &PSPNet& B  & \textbf{74.4}   & \textbf{75.6} & \textbf{73.9}  & \textbf{74.9} \\
		\emph{$A^2$GNN} (ours)& -  &Tree-FCN~\cite{song2019learnable}& B  & \textbf{75.1}   & \textbf{76.5} & \textbf{74.5}  & \textbf{75.2} \\ \hline                                     
	\end{tabular}\label{tab:SOTA}
\end{table*}
\section{Implement Details}

To generate pixel-level seed label from image-level label, we use the same classification network as SEAM~\cite{wang2020self}, which is a ResNet-38~\cite{wu2019wider}. All the parameters are kept the same as in~\cite{wang2020self}. 

Our affinity CNN adopts the same backbone with the above classification network. At the same time, dilated convolution is used in the last three residual blocks and their dilated rates are set as $2$, $4$ and $4$, respectively. As in Fig.~\ref{fig:AffinityNet}, the output channels of these three residual blocks are $512$, $1024$ and $4096$. A node feature is a concatenated feature of these three outputs, so the feature dimension for one node is $5632$. Since we need to use feature to compute distance, three $1 \times 1$ convolution kernels are used to reduce the feature dimensions of these three residual blocks and the output channels are set as $64$, $128$ and $256$, respectively. Finally, a $1 \times 1$ convolution kernel with $448$ channels is used to get the final feature map $F_A$. Following \cite{ahn2018learning}, we set $r=5$ for both training and inference. $\lambda$ in Eq.~(\ref{L_A}) is set to $3$ and $\sigma_{xy} = 6$, $\sigma_{rgb}=0.1$.

Our \emph{$A^2$GNN} has five layers as mentioned in section~\ref{sec:AAL}, the output channel number for the first layer and three affinity attention layers are $256$. $\lambda_1$ in Eq.~(\ref{eq:L_G}) are set as 0.01. In $\mathcal{L}_{reg}$, we adopt the same parameters with Eq.~(\ref{Kab}). We use Adam as optimizer~\cite{kingma2014adam} with the learning rate being $0.03$ and weight decay being $5\times10^{-4}$. During training, the epoch number is $100$ and the dropout rate is $0.5$. The training process will be divided into two stages: In the first stage (the first 50 epochs), $L_{reg}$ and consistency-checking are not used while in the second stage, all losses and consistency-checking are used. We use dropout after the first layer. We use bilinear interpolation to achieve the original resolution during training and inference. CRF \cite{krahenbuhl2013parameter} is used as the post-processing method during inference. The unary potential of CRF uses the final output probability $O$ in Eq.~(\ref{AO}) while pair-wise potential corresponds to the color and spatial position of different nodes. All CRF parameters are the same as~\cite{ahn2018learning,zhang2019reliability}. Note that for the BSIS task, we need to convert the above pseudo labels to instance masks. Given a bounding box, we directly assign pixels which locate inside a bounding box and share the same class with it to one instance.

For the BSSS task, we take the Deeplab-Resnet101~\cite{chen2017deeplab}, PSPNet~\cite{zhao2017pspnet} and Tree-FCN~\cite{song2019learnable} as our fully supervised semantic segmentation models for fair comparison. For the BSIS task, MaskR-CNN~\cite{he2017mask} is taken as the final instance segmentation model and we use Resnet-101 as the backbone. Following the same post-processing with \cite{hsu2019weakly}, we use CRF~\cite{krahenbuhl2013parameter} to refine our final prediction.

All experiments are run on 4 Nvidia-TiTan X GPUs. For Pascal VOC 2012 dataset, generating the pixel-level seed label takes about 12 hours, training affinity CNN spends about 12 hours and generating the pseudo labels using \emph{$A^2$GNN} takes about 16 hours.   

\section{Experiment}
\subsection{Datasets}
We evaluate our method on PASCAL VOC 2012 \cite{everingham2011pascal} and COCO~\cite{lin2014microsoft} dataset. For PASCAL VOC 2012, the augmented data SBD~\cite{6126343} is also used, and the whole dataset includes 10,582 images for training and 1,449 images for validating and 1,456 images for testing. For COCO dataset, we train on the default train split (80K images) and then test on the test-dev set.

For Pascal VOC 2012 dataset, mean intersection over union (mIoU) is applied as the evaluation criterion for weakly supervised semantic segmentation, and the mean average precision (mAP) \cite{hariharan2014simultaneous} is adopted for weakly supervised instance segmentation. Following the same evaluation protocol as prior works, we reported mAP with three thresholds (0.5, 0.7, 0.75), denoting as $\text{mAP}_{0.5}^r$, $\text{mAP}_{0.7}^r$ and $\text{mAP}_{0.75}^r$, respectively. For COCO dataset, following~\cite{fan2018associating}, mAP, $\text{mAP}_{0.5}^r$, $\text{mAP}_{0.75}^r$, $\text{mAP}_{s}$, $\text{mAP}_{m}$ and $\text{mAP}_{l}$ are reported. 

\subsection{Comparison with State-of-the-Art}

\textbf{Weakly supervised semantic segmentation:} In Table~\ref{tab:SOTA}, we compare the performance between our method and other state-of-the-art approaches for BSSS. For using deeplab as the segmentation model, it can be seen that our approach obtains 96.1\% of our upper-bound with pixel-level supervision (Deeplab-Resnet101~\cite{chen2017deeplab} with CRF). Compared to the other approaches, our approach gives a new state-of-the-art performance. Specifically, our approach with deeplab-resnet101~\cite{song2019learnable} outperforms Box-Sup~\cite{dai2015boxsup}, WSSL~\cite{papandreou1502weakly} by big margins, approximately 11.8\% and 13.2\%, respectively. Besides, compared to \emph{GraphNet}~\cite{pu2018graphnet}, the only graph learning solution, our method with Deeplab-Resnet101 performs much better than it, with an improvement of 10.9\% for mIoU (without CRF). We can also observe that our performance is even better than SDI \cite{khoreva2017simple}, which uses MCG \cite{arbelaez2014multiscale} and BSDS \cite{martin2001database} as extra pixel-level supervision. When using PSPNet~\cite{zhao2017pspnet} as the segmentation model, our approach obtains 74.4\% mIoU without CRF as post-processing, which is even higher than the results in~\cite{li2018weakly} with CRF. Finally, our method with Tree-FCN~\cite{song2019learnable} outperforms the state-of-the-art Box2Seg~\cite{kulharia2020box2seg} in this task. Note that Box2Seg focused on designing a segmentation network using noisy label from bounding box, thus our performance could be further improved using their network as the final segmentation network.   

\begin{table}[t]
	\caption{Comparison with other approaches on PASCAL VOC 2012 \emph{val} dataset for BSIS.}
	\centering
	\resizebox{\columnwidth}{!}{
		\begin{tabular}{lccccc}
			\hline
			Method    & Pub. & Sup. & $\text{mAP}_{0.5}^r$ & $\text{mAP}_{0.7}^r$ & $\text{mAP}_{0.75}^r$ \\ \hline
			SDS~\cite{hariharan2014simultaneous} & ECCV'14 & F    & 49.7   & 25.3   & -   \\
			MaskR-CNN~\cite{he2017mask} & ICCV'17 & F    & 67.9   & 52.5   & 44.9    \\ \hline
			PRM~\cite{zhou2018weakly} & CVPR'18 & I    & 26.8   & -   & 9.0    \\
			IRN~\cite{ahn2019weakly} & CVPR'19 & I    & 46.7   & 23.5   & -    \\
			SDI~\cite{khoreva2017simple}       & CVPR'17 & B    & 44.8   & -      & 16.3    \\
			BBTP~\cite{hsu2019weakly}      & NeurIPS'19 & B    & 58.9   & 30.4   & 21.6    \\
			\emph{$A^2$GNN} (ours)      & -            & B    & \textbf{59.1}       &  \textbf{35.5}      & \textbf{27.4}    \\ \hline
		\end{tabular}
	}
	\label{tab:InsatnceSOTA}
\end{table}

\begin{table*}[]
	\centering
	\caption{Comparison with other approaches on COCO test-dev dataset for weakly supervised instance segmentation. E: extra dataset~\cite{li2017instance} with instance-level annotation. S$^4$Net: salient instance segmentation model~\cite{fan2019s4net}.}
	\begin{tabular}{lcccccccc}
		\hline
		Method & Pub. & Sup. & mAP & $\text{mAP}_{0.5}^r$ & $\text{mAP}_{0.75}^r$ & AP$_s$& AP$_m$ & AP$_l$ \\ \hline
		MNC~\cite{dai2016instance}&CVPR'16&F&24.6&44.3& 24.8&4.7        &25.9&43.6\\
		Mask-RCNN~\cite{he2017mask}&ICCV'17&F&37.1&60.0&39.6&35.3 &35.3&35.3     \\
		\hline
		Fan \emph{et.al.}~\cite{fan2018associating}&ECCV'18& I+E+S$^4$Net&13.7&25.5&13.5&0.7&15.7&26.1\\
		LIID~\cite{liu2020leveraging}&PAMI'20& I+E+S$^4$Net&16.0&27.1&16.5&3.5&15.9&27.7\\
		\emph{$A^2$GNN} (ours)&-&B&\textbf{20.9}&\textbf{43.9}&\textbf{17.8}&\textbf{8.3}&\textbf{20.1}&\textbf{31.8}\\ \hline
	\end{tabular}\label{tab:COCO}
\end{table*}

\begin{figure*}
	\centering
	\includegraphics[width=\textwidth]{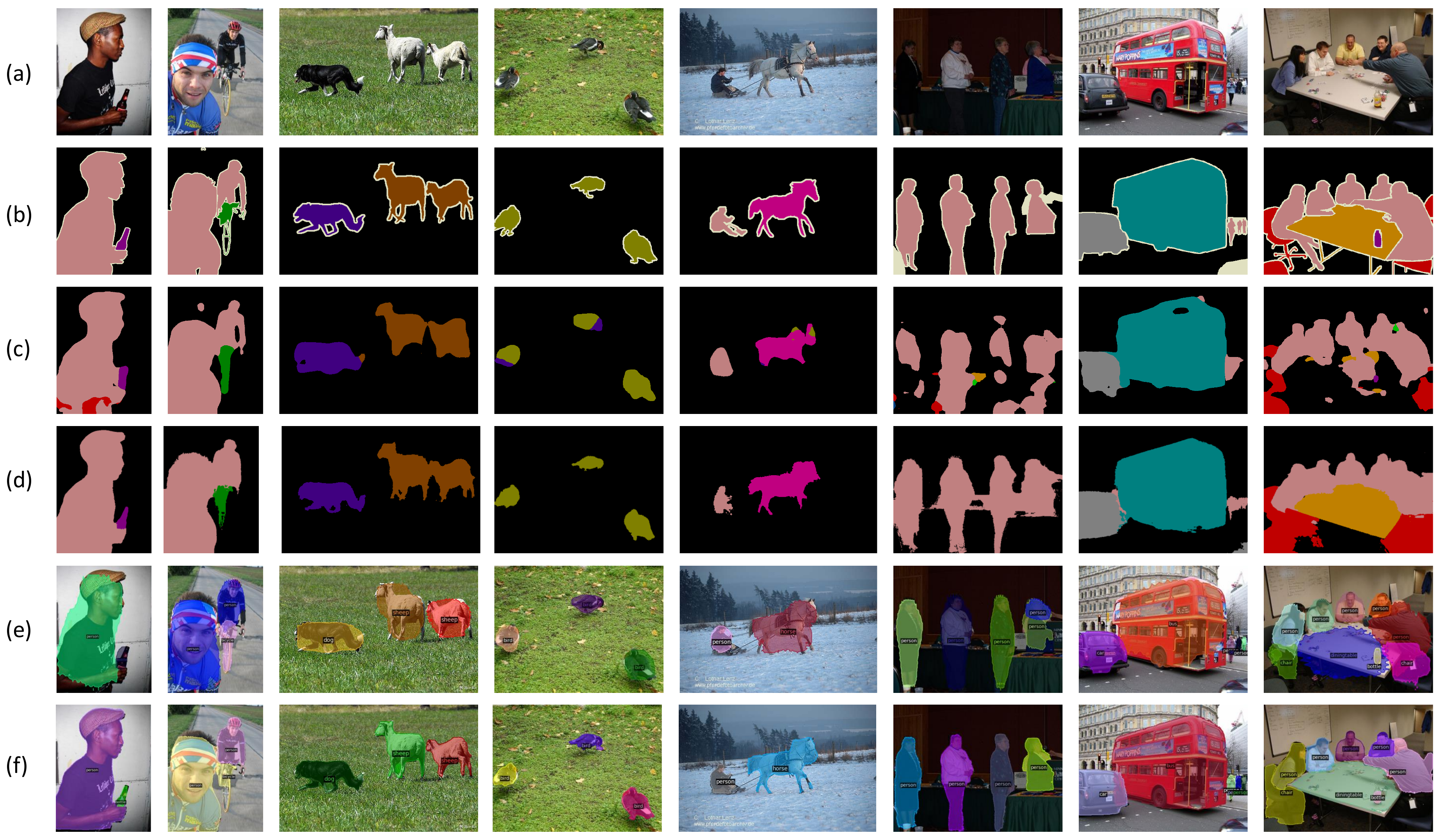}
	\caption{Qualitative results of our \emph{$A^2$GNN} and other state-of-the-art approaches on PASCAL VOC 2012 \emph{val} dataset. (a) Original image. (b) Ground truth of semantic segmentation. (c) SDI~\cite{khoreva2017simple} for BSSS. (d) Our results for BSSS. (e) BBTP~\cite{hsu2019weakly} for BSIS. (f) Our results for BSIS.}
	\label{fig:vis}
\end{figure*} 

\textbf{Weakly supervised instance segmentation:} In Table~\ref{tab:InsatnceSOTA}, we compare our approach to other state-of-the-art approaches on BSIS. It can be seen that our approach achieves a new state-of-the-art performance among all evaluation criteria. Specifically, our approach performs much better than SDI~\cite{khoreva2017simple}, increasing 14.3\% and 11.1\% on $\text{mAP}_{0.5}^r$ and $\text{mAP}_{0.75}^r$, respectively. It can also be found that compared to BBTP~\cite{hsu2019weakly}, which is the state-of-the-art approach on this task, our approach significantly outperforms it by large margins, around 5.1\% on $\text{mAP}_{0.7}^r$ and 5.8\% on $\text{mAP}_{0.75}^r$. The performance is increased more on $\text{mAP}_{0.75}^r$ than $\text{mAP}_{0.7}^r$ and $\text{mAP}_{0.5}^r$, which also indicates that our approach can produce masks that preserve the object structure details. One interesting observation is that our approach even achieves better performance than the fully supervised method SDS~\cite{hariharan2014simultaneous}. 

In Fig.~\ref{fig:vis}, we compare some qualitative results between our approach and other state-of-the-art approaches for which the source code is publicly available. Specifically, we compare our results with SDI~\cite{khoreva2017simple}\footnote[1]{we use a re-implement code from: \href{https://github.com/johnnylu305/Simple-does-it-weakly-supervised-instance-and-semantic-segmentation}{github.com/johnnylu305}} for the BSSS task and BBTP~\cite{hsu2019weakly}\footnote[2]{\href{https://github.com/chengchunhsu/WSIS_BBTP}{github.com/chengchunhsu/WSIS\_BBTP}} for the BSIS task. It can be seen that compared to other approaches, our approach produces better segmentation masks covering object details.

In Table~\ref{tab:COCO}, we make a comparison between our approach and others on COCO test-dev dataset. It can be seen that our approach performs much better than LIID~\cite{liu2020leveraging}, with an increase of 16.8\% on mAP$^r_{\text{50}}$. Furthermore, our approach even performs competitive with fully-supervised approach MNC~\cite{dai2016instance}, which also indicates the effectiveness of our approach.

\begin{figure}[!htb]
	\centering
	\includegraphics[width=\columnwidth]{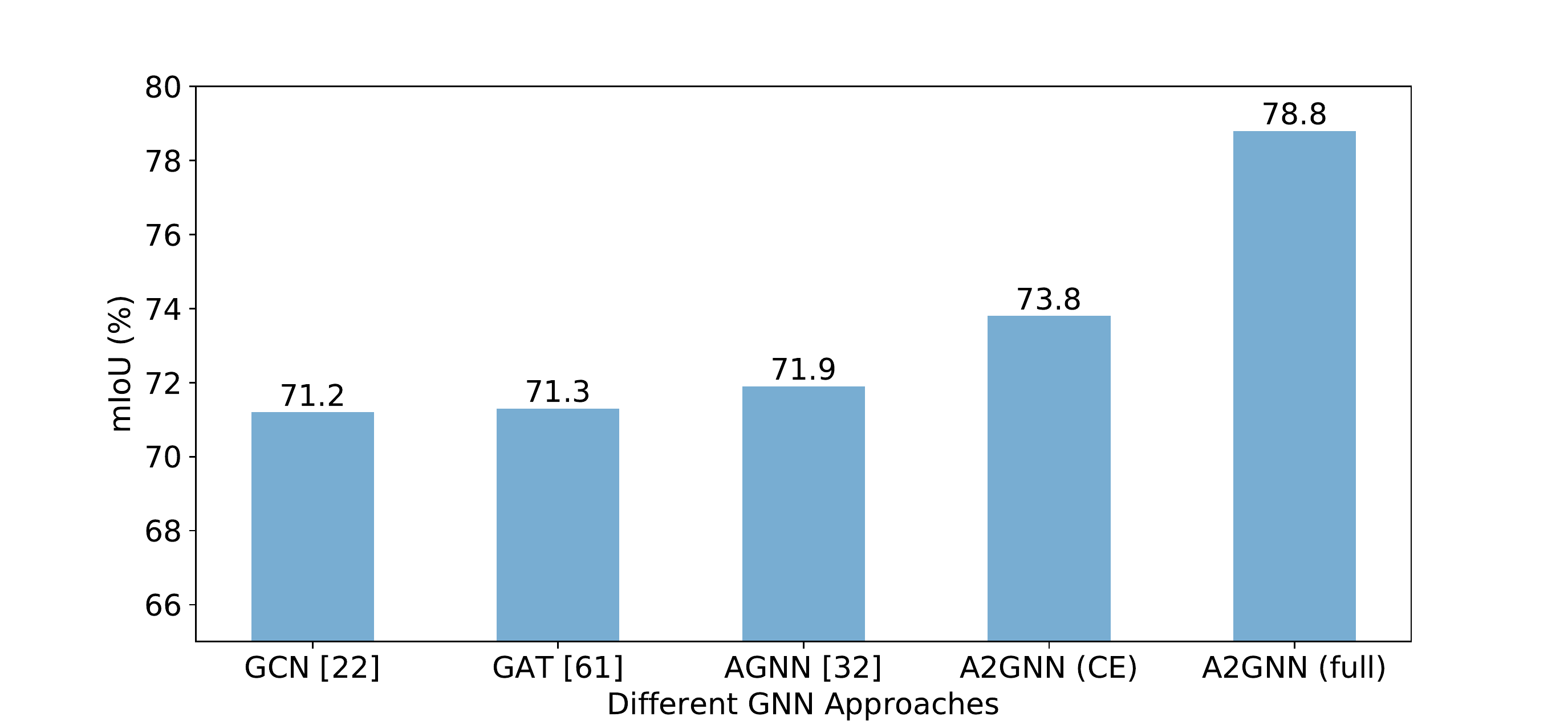}
	\caption{Comparison between our \emph{$A^2$GNN} and other GNNs (GCN~\cite{kipf2016semi}, GAT~\cite{velivckovic2017graph}, AGNN~\cite{guo2019attention}) on Pascal VOC 2012 \emph{training} set. ``CE" means only cross-entropy loss is used.}
	\label{fig:VsGNNs}
\end{figure}

\begin{table*}[t]
	\caption{Ablation studies on Pascal VOC 2012 \emph{training} dataset for BSSS.}
	\centering
		\subtable[Evaluation for different modules in our approach. \emph{RW}:random walk~\cite{wang2020self}. \emph{H}: affinity attention layer. \emph{C.C.}: consistency-checking.]{
			\resizebox{0.45\textwidth}{!}{
				\begin{tabular}{ccccccccc}
					\hline
					\multicolumn{1}{l}{\multirow{2}{*}{Baseline}} & \multicolumn{2}{c}{affinity CNN}& \multirow{2}{*}{\emph{RW}} & \multicolumn{4}{c}{\emph{$A^2$GNN}}                  & \multirow{2}{*}{mIoU} \\ \cline{2-3} \cline{5-8}
					\multicolumn{1}{l}{}& \multicolumn{1}{l}{$\mathcal{L}_{Ac}$} & \multicolumn{1}{c}{$\mathcal{L}_{Ar}$} &  & \multicolumn{1}{c}{\emph{H}} & \multicolumn{1}{l}{$\mathcal{L}_{reg}$} & \multicolumn{1}{l}{$\mathcal{L}_{mp}$} & \multicolumn{1}{l}{\emph{C.C.}} &  \\ \hline
					\checkmark  & & & & & & &&62.3\\
					\checkmark  & \checkmark & &\checkmark &  & & &&70.3\\
					\checkmark  & \checkmark & \checkmark & \checkmark  & & & &  &71.3\\
					\checkmark   & \checkmark & \checkmark  &  & \checkmark    &   &  &   &73.8\\
					\checkmark & \checkmark  & \checkmark  &  & \checkmark   & \checkmark &  & &74.9\\
					\checkmark  & \checkmark  & \checkmark &  & \checkmark   & \checkmark & \checkmark  &  &78.1\\
					\checkmark & \checkmark  & \checkmark  &   & \checkmark  & \checkmark  & \checkmark  & \checkmark &\textbf{78.8}\\ \hline
				\end{tabular}\label{tab:Baseline}
			}
		}
	\qquad
	\subtable[Evaluation for the loss functions of our \emph{$A^2$GNN}. \emph{C.C.}: consistency-checking.]{       
		\resizebox{0.35\textwidth}{!}{
			\begin{tabular}{ccccc}
				\hline
				$\mathcal{L}_{ce}$ & $\mathcal{L}_{reg}$ & $\mathcal{L}_{mp}$ & \emph{C.C.}& mIoU (\%) \\ \hline
				\checkmark&     &    & &73.8\\
				\checkmark&\checkmark&    & &74.9 \\
				\checkmark&     &\checkmark& &75.9\\
				\checkmark&     &\checkmark& \checkmark&77.2\\
				\checkmark&\checkmark&\checkmark& &78.1\\
				\checkmark&\checkmark &\checkmark&\checkmark&\textbf{78.8}\\ \hline
			\end{tabular}\label{tab:GNet_loss}
		}
	}
			
	\subtable[Evaluation for different methods to build the graph. S.P.: superpixel. Feat.: feature map. Dis.: distance. Aff: affinity CNN.]{       
		\resizebox{0.31\textwidth}{!}{
			\begin{tabular}{ccccc}
			\hline
			S.P. & Feat.& Dis. & Aff. & mIoU(\%) \\ \hline
			\checkmark&  &\checkmark&   &73.3\\
			&\checkmark&\checkmark& &74.7\\
			& \checkmark& &\checkmark& \textbf{78.8}\\ \hline
			\end{tabular}
			}
			\label{tab:AffinityNet}
	}	
 	\qquad
	\subtable[Evaluation of the affinity attention layer in \emph{$A^2$GNN}.]{
		\resizebox{0.18\textwidth}{!}{
	\begin{tabular}{ccc}
		\hline
		\multicolumn{2}{c}{\emph{$A^2$GNN} layer} & \multirow{2}{*}{mIoU(\%)} \\ \cline{1-2}
		cos($\cdot$)& edge &\\ \hline
		\checkmark &  &73.1\\
		&\checkmark &77.3\\
		\checkmark&\checkmark&\textbf{78.8}\\ \hline
	\end{tabular} \label{tab:AGNN}
		}
	}
 	\qquad
	\subtable[Performance comparison for using different seed labels on affinity CNN and the loss functions.]{
		\resizebox{0.25\textwidth}{!}{
		\begin{tabular}{ccccc}
			\hline
			$\mathcal{L}_{Ac}$ & $\mathcal{L}_{Ar}$&$M_F$ &$M_g$  &mIoU\\ \hline
			\checkmark& &\checkmark&&74.5\\
			\checkmark&  &  &\checkmark&71.7\\
			\checkmark&\checkmark&\checkmark&   &76.1\\
			\checkmark&\checkmark&&\checkmark&\textbf{78.8}\\ \hline
		\end{tabular}\label{tab:Aff_label_loss}
		}
	}
 	\qquad
			
\end{table*}

\subsection{Ablation Studies}
Since the pseudo labels for BSIS are generated from the BSSS task, in this section, we will conduct ablation studies only on the BSSS task. We simply evaluate the pseudo label mIoU on the \emph{training} set, without touching the \emph{val} and \emph{test} set. 

In Fig~\ref{fig:VsGNNs}, we make a comparison between our \emph{$A^2$GNN} and others for BSSS. It can be seen that our \emph{$A^2$GNN} performs much better than other GNNs, with an improvement of 1.9\% mIoU over AGNN~\cite{guo2019attention} when only using the cross-entropy loss, and the full \emph{$A^2$GNN} outperforms AGNN~\cite{guo2019attention} by a large margin (6.9\%).

In Table~\ref{tab:Baseline}, we explore the influence of different modules in our approaches to generate pseudo labels. \emph{Baseline} means that we use SEAM~\cite{wang2020self} to generate the foreground seed labels and then use bounding box supervision to generate the background. \emph{RW} means that we follow SEAM~\cite{wang2020self} to use random walk for pseudo label generation. It can be seen that the proposed approach outperforms the baseline by a large margin. And each module significantly improves the performance. 

In Table~\ref{tab:GNet_loss}, we study the effectiveness of our joint loss function. It can be seen that compared to \emph{$A^2$GNN} which only adopts cross-entropy loss, our MP loss can improve its performance by 2.1\%, validating the effectiveness of our MP loss. With consistency-checking, the performance is improved to 77.2\%, indicating the effectiveness of our proposed consistency-checking mechanism. When jointly optimized by these three losses with our consistency-checking mechanism, the performance is further improved to 78.8\%.

In Table~\ref{tab:AffinityNet}, we study different ways to build our graph.
\emph{Superpixel (S.P.)} means that we adopt \cite{pu2018graphnet} to produce graph nodes and their features. \emph{Distance (Dis.)} means that we build the graph edge using  $L_1$ distance of feature map \cite{pu2018graphnet}. It can be seen that the performance is improved when directly using pixel in the feature map as the node, suggesting that it is more accurate than using superpixel. When we use our affinity CNN to build the graph, the performance is significantly improved by 4.1\%, which shows that our approach can build a more accurate graph than other approaches. 

In Table~\ref{tab:AGNN}, we study the effectiveness of our affinity attention layer. It can be found that if we use either the attention module or the affinity module separately, the mIoU score is lower than that of the full \emph{$A^2$GNN}, which indicates the effectiveness of our designed GNN layer.

In Table~\ref{tab:Aff_label_loss}, we show the joint influence of the loss functions and labels for our proposed affinity CNN. It can be seen that when only using $L_{Ac}$, $M_F$ labels perform better than $M_g$. This is because that $M_g$ only provides limited pixels and these pixels are usually located at the discriminative part of an object (such as the human head). Such limited labels are not sufficient when only using $L_{Ac}$. When we use both $L_{Ac}$ and $L_{Ar}$, $M_g$ performs much better than $M_F$, indicating that $L_{Ar}$ can accurately propagate the labeled regions to unlabeled regions.

In addition, we also analyze the influence of supervision for our \emph{$A^2$GNN}. Specifically, we make a comparison of the results when using $M_F$ (in Eq.~(\ref{MF})) and $M_g$ (in Eq.~(\ref{LVl})) as supervision for our \emph{$A^2$GNN}, respectively. Compared to $M_F$, $M_g$ has fewer annotated nodes but each annotation is more reliable. The mIoU score on Pascal VOC 2012 \emph{training} set is 73.2\% and 78.8\% for $M_F$ and $M_g$, respectively. This result validates the effectiveness of the leverage of the high-confident labels.
	\begin{figure*}[!htb]
		\centering
		\includegraphics[width=\textwidth]{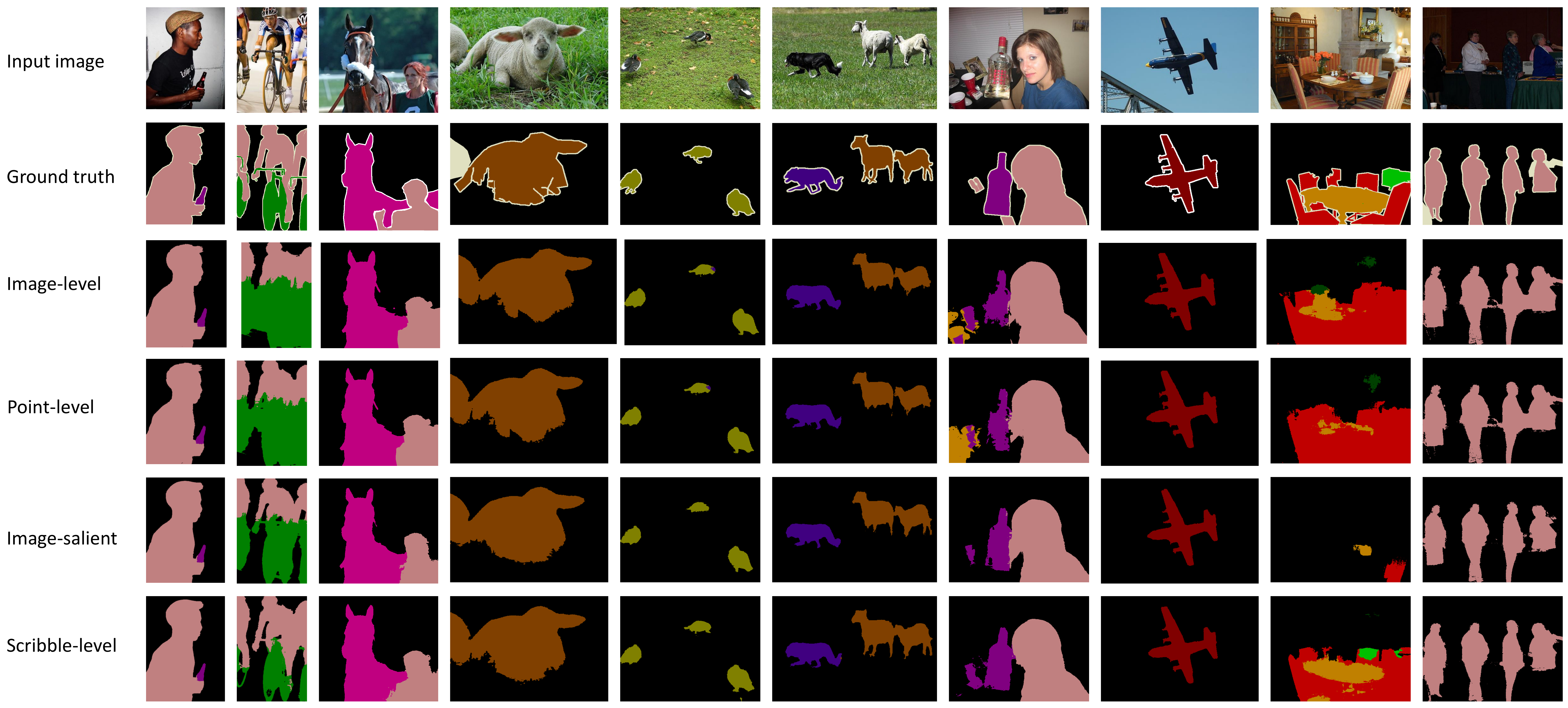}
		\caption{Qualitative results of our $A^2$GNN on PASCAL VOC 2012 \emph{val} dataset. We show the results from different levels of supervision signals (3rd -- 6th rows). Stronger supervision signals (\eg, scribble)  produce more accurate results than weaker signals (\eg, point, image-level label).}
		\label{fig:vis_appendix}
	\end{figure*} 

\section{Application to Other Weakly Supervised Semantic Segmentation Tasks}
	In order to use our approach on other weakly supervised semantic segmentation tasks, \eg, scribble, point and image-level, we need to ignore our proposed MP loss (section~\ref{sec:MP}) and the consistency-checking (section~\ref{sec:cc}) as they rely on bounding box supervision.  Besides, we need to convert different weak supervised signals to pixel-level seed labels. All other steps and parameters are the same as that in the BSSS task. In the following section, we will introduce how to convert the different weakly supervised signal to pixel-level seed labels, and then we will report experimental results on these tasks. 
		\begin{table}[t]
			\centering
			\caption{Comparison with other state-of-the-arts on PASCAL VOC 2012 \emph{val} and \emph{test} datasets. Sup.: Segmentation model. F: fully supervised. S: scribble. P: point. I: image-level label. E: extra salient dataset. ``highlight" means the best performance for a specific task.}
			\begin{threeparttable}
				\begin{tabular}{lcccccccc}
					\hline
					Method & Pub. & Seg. &Sup. & \emph{val} & \emph{test} \\ \hline
					(1) Deeplab-V1~\cite{chen2014semantic} &- &-& F  & 67.6  & 70.3 \\
					(2) Deeplab-Vgg \cite{chen2017deeplab} &  PAMI'18&-& F  & 71.5  & 72.6 \\
					(3) Deeplab-Resnet~\cite{chen2017deeplab} & PAMI'18 &-& F  & 76.8 & 79.7 \\
					(4) WiderResnet38 \cite{wu2019wider} & PR'19 &-&F & 80.8   & 82.5 \\
					(5) Tree-FCN~\cite{song2019learnable} & NeurIPS'19  &-&F & 80.9   & -\\ \hline 
					RAWKS  \cite{vernaza2017learning} & CVPR'17 &(1)&S  & 61.4  & - \\
					ScribbleSup  \cite{lin2016scribblesup} & CVPR'16 &(2)&S  & 63.1  & - \\
					GraphNet~\cite{pu2018graphnet} & ACMM'18 &(3)&S  & 73.0 &  - \\
					Regularized loss  \cite{tang2018regularized} & ECCV'18 &(3)&S  &75.0 & - \\
					\emph{$A^2$GNN} (ours) & - &(3)&S  & 74.3 & 74.0  \\
					\emph{$A^2$GNN} (ours) & - &(5)&S  & \textbf{76.2} & \textbf{76.1}  \\ \hline
					What's the point  \cite{bearman2016s} & ECCV'16 &(2)&P & 43.4   & 43.6 \\
					Regularized loss  \cite{tang2018regularized} & ECCV'18 &(3)&P  & 57.0 &-\\
					\emph{$A^2$GNN}(ours) & - &(3)&P & \textbf{66.8} & \textbf{67.7} \\ \hline
					AE-PSL  \cite{wei2017object} & CVPR'17 &(1)&I+E  &55.0 & 55.7 \\
					DSRG  \cite{huang2018weakly} & CVPR'18 &(3)&I+E   & 61.4 &  63.2 \\
					FickleNet  \cite{lee2019ficklenet} & CVPR'19 &(3)&I+E & 64.9 & 65.3 \\
					Zhang \emph{et.al}~\cite{zhang2020splitting}& ECCV'20 &(3)&I+E & 66.6 & 66.7 \\
					ICD \cite{fan2020learning} & CVPR'20 &(3)&I+E & 67.8 & 68.0 \\
					EME~\cite{fan2020employing}& ECCV'20 &(3)&I+E & 67.2 & 66.7 \\
					MCIS~\cite{sun2020mining}& ECCV'20 &(3)&I+E & 66.2 & 66.9 \\
					ILLD~\cite{liu2020leveraging}& PAMI'20 &(3)&I+E & 66.5 & 67.5 \\
					ILLD~\cite{liu2020leveraging}& PAMI'20 &(3)$^\dag$&I+E & \textbf{69.4} & \textbf{70.4} \\
					\emph{$A^2$GNN}(ours) & - &(3)&I+E & 68.3  & 68.7 \\ 
					\emph{$A^2$GNN}(ours) & - &(3)$^\dag$&I+E & 69.0 & 69.6 \\ \hline
					PSA \cite{ahn2018learning}& CVPR'18&(4)&I & 61.7  & 63.7 \\
					SEAM \cite{wang2020self}& CVPR'20 &(4) &I & 64.5 & 65.7 \\
					ICD \cite{fan2020learning}& CVPR'20 &(3)&I & 64.1 & 64.3 \\
					BES~\cite{chen2020weakly} & ECCV'20 &(3)&I & 65.7 & 66.6 \\
					SubCat~\cite{chang2020weakly} & CVPR'20 &(3)&I & 66.1 & 65.9 \\
					CONTA~\cite{zhang2020causal} & NeurIPS'20 &(4)&I & 66.1 & 66.7 \\
					\emph{$A^2$GNN}(ours) & - &(3)&I & \textbf{66.8} & \textbf{67.4} \\
					\hline
				\end{tabular}
				\begin{tablenotes}
					\item[\dag] means using Res2Net~\cite{gao2019res2net} as the backbone.
				\end{tablenotes}
			\end{threeparttable}
			\label{SOTA}
		\end{table}
		
	\subsection{Pixel-level Seed Label Generation}	
	As mentioned in section~\ref{sec:CAM}, the common practice to initialize weakly supervised task is to generate pixel-level seed labels from the given weak supervision. For different weakly supervision, we use different approaches to convert them to pixel-level seed labels.
	
	\textbf{Image-level supervision:} we directly use $M'_{I}$ defined in Eq.~(\ref{LVl}) to train our affinity CNN and use it as $M_g$ to train our \emph{$A^2$GNN}. The final pseudo labels are generated using the ratio (1:3) to fuse our results and the results of random walk.
	
	\textbf{Scribble supervision:} For the scribble supervised semantic segmentation task, for each class in an image (including background), it provides one or more scribbles as labels. Superpixel method~\cite{achanta2012slic} is used to get the expanded labels $M_{S}$ from the initial scribbles.
	To get seed label to train our affinity CNN, we merge $M_{S}$ with $M'_{I}$ using the following rule: if the pixel label in $M_S$ is known (not 255), the corresponding label in $M_g$ will be the same label as $M_{S}$. Otherwise, the pixel label will be treated as the same label as $M'_{I}$. To generate the node labels for \emph{$A^2$GNN}, we directly use $M_{g} = M_{S}$ since it provides accurate labels for around 10\% pixels in an image.
	
	\textbf{Point supervision:} For point supervised semantic segmentation, for each object in an image, it provides one point as supervision and there is no annotation for background. To train our affinity CNN, we used $M'_{I}$ directly. To generate node supervision for \emph{$A^2$GNN}, we use a superpixel method \cite{achanta2012slic} to get the expanded label $M_{P}$ from initial point labels. Then $M_g$ is generated using the same setting with the scribble task.
	
	For our affinity CNN and our \emph{$A^2$GNN}, we use the same setting with our bounding box task. 
	
	\subsection{Experimental Evaluations}
	In Table \ref{SOTA}, we compare the performance between our method and other state-of-the-art weakly supervised semantic segmentation approaches. 
	
	For point supervision, our method achieves state-of-the-art performance with 66.8\% and 67.7\% mIoU on the \emph{val} and \emph{test} set of PASCAL VOC, respectively. Compared to other two approaches~\cite{bearman2016s}  and~\cite{tang2018regularized}, our method increases 23.4\% and 9.8\% in mIoU on PASCAL VOC 2012 \emph{val} dataset, respectively.
	
	\begin{table}
		\centering
		\caption{Performance comparison in mIoU (\%) for evaluating the pseudo labels on the PASCAL VOC training data set.}
		\begin{threeparttable}
			\begin{tabular}{cccc}
				\hline
				Method & Pub. & Sup. & mIoU (\%) \\ \hline
				PSA~\cite{ahn2018learning}    & CVPR'18 &I& 58.4\\
				ICD~\cite{fan2020learning}    & CVPR'20 &I& 62.2\\
				SubCat~~\cite{chang2020weakly} & CVPR'20 &I & 63.4 \\
				SEAM~\cite{wang2020self}   & CVPR'20 &I & 63.6 \\
				\emph{$A^2$GNN} (ours)  & - &I& \textbf{65.3} \\
				\emph{$A^2$GNN} (ours)  & - &I+E& \textbf{66.5} \\ \hline
				Box2Seg~\cite{kulharia2020box2seg} & ECCV'20 &B& 73.6\tnote{*} \\
				\emph{$A^2$GNN} (ours)  & - &B& \textbf{78.8} \\ \hline
			\end{tabular}
			\begin{tablenotes}
				\item[*] Reproduce by ourself.
			\end{tablenotes}
		\end{threeparttable}
		\label{tab:pseudo_label}
	\end{table}	
		
	For the image-level supervision task, our \emph{$A^2$GNN} achieves mIoU of 66.8\% and 67.4\% on \emph{val} and \emph{test} set, respectively. It should be noticed that PSA~\cite{ahn2018learning}, SEAM~\cite{wang2020self} and CONTA~\cite{zhang2020causal} apply Wider ResNet-38~\cite{wu2019wider} as segmentation model, which has a higher upper-bound than Deeplab-Resnet101~\cite{chen2017deeplab}.  Using Deeplab-Resnet101~\cite{chen2017deeplab} as the segmentation, Subcat~\cite{chang2020weakly} is the state-of-the-art approach on this task, but it require multi-round training processes. Moreover, our method achieves 66.8\% mIoU using Deeplab-Resnet101~\cite{chen2017deeplab}, being 87.0\% of our upper-bound (76.8\% mIoU score with Deeplab-Resnet~\cite{chen2017deeplab}) on \emph{val} set.	
	
	Besides, for the image-level supervision, some approaches~\cite{huang2018weakly,lee2019ficklenet,fan2020learning, liu2020leveraging} used salient model with extra pixel-level salient dataset~\cite{hou2017deeply} or instance pixel-level salient dataset~\cite{li2017instance} to generate more accurate pseudo labels. Follow these approaches, we also use saliency models. Specifically, we use the saliency approach~\cite{liu2019simple} following ICD~\cite{fan2020learning} to produce the initial seed labels, and then use our approach to produce the final pseudo labels. It can be seen from Table~\ref{SOTA} that our approach outperforms other approaches (using ResNet101 as the backbone).Following ILLD~\cite{liu2020leveraging}, we also evaluate our approach using Res2Net~\cite{gao2019res2net} as the segmentation backbone, and our performance is further improved to 69.0\% and 69.6\%. For this setting, we have not designed any specific denoising scheme for the seed labels. Nevertheless, our performance is comparable with other state-of-the-art methods, \emph{e.g.,}~\cite{liu2020leveraging}, which also proves that our method can be well generalized to all weakly supervised tasks.
	
	For the scribble supervision task, our method also achieves a new state-of-the-art performance. 
	
	In Table~\ref{tab:pseudo_label}, we present a comparison to evaluate the pseudo labels on the PASCAL VOC training set. It can be seen that our approach outperforms other approaches. Compared to the state-of-the-art approach SEAM~\cite{wang2020self}, our approach obtains 1.7\% mIoU improvement. We also compare the quality of the pseudo labels between our approach and Box2Seg~\cite{kulharia2020box2seg}. It can be seen that our method outperforms Box2Seg~\cite{kulharia2020box2seg} by a large margin, with 5.2\% mIoU improvement.
	
	In Fig.~\ref{fig:vis_appendix}, we also present more qualitative results for the above three tasks. It can be seen that stronger supervision leads to better performance and preserves more segmentation details. 
	
\section{Conclusion}
We have proposed a new system, \emph{$A^2$GNN}, for the bounding box supervised semantic segmentation task. With our proposed affinity attention layer, features can be accurately aggregated even when noise exists in the input graph. Besides, to mitigate the label scarcity issue, we further proposed a MP loss and a consistency-checking mechanism to provide more reliable guidance for model optimization. Extensive experiments show the effectiveness of our proposed approach. In addition, the proposed approach can also be applied to bounding box supervised instance segmentation and other weakly supervised semantic segmentation tasks. As future work, we will investigate how to generate more reliable seed labels and more accurate graph, so that the noise level in the input graph can be alleviated and therefore our \emph{$A^2$GNN} can produce more accurate pseudo labels. 

\bibliographystyle{IEEEtran}
\bibliography{mybib}


%

\begin{IEEEbiography}[{\includegraphics[width=1in,height=1.25in,clip,keepaspectratio]{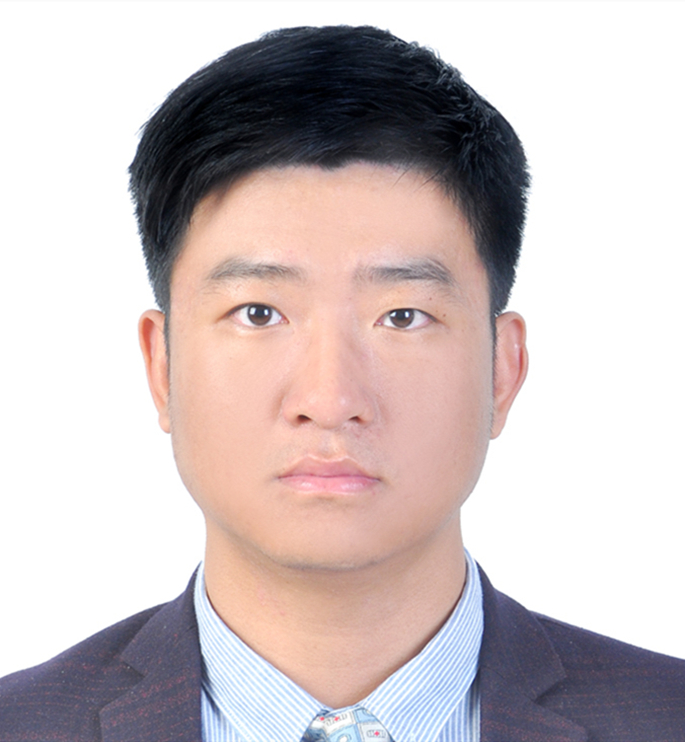}}]{Bingfeng Zhang}
	received the B.S. degree in electronic information engineering from China University of Petroleum (East China), Qingdao, PR China, in 2015, the M.E. degree in systems, control and signal processing from University of Southampton, Southampton, U.K., in 2016. He is now a Ph.D student in the University of Liverpool, Liverpool, U.K., and also a Ph.D student in the school of the advanced technology of the Xi'an Jiaotong-Liverpool University, Suzhou, PR China. His current research interest is weakly supervised semantic segmentation and few-shot segmentation. 
\end{IEEEbiography}

\begin{IEEEbiography}[{\includegraphics[width=1in,height=1.25in,clip,keepaspectratio]{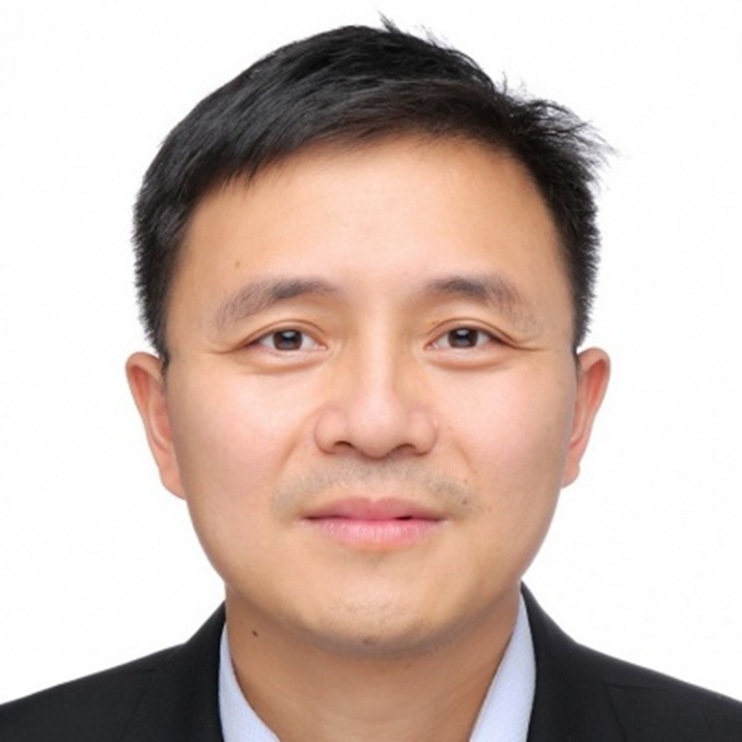}}]{Jimin Xiao}
	received the B.S. and M.E. degrees in telecommunication engineering from the Nanjing University of Posts and Telecommunications, Nanjing, China, in 2004 and 2007, respectively, and the Ph.D. degree in electrical engineering and electronics from the University of Liverpool, Liverpool, U.K., in 2013. From 2013 to 2014, he was a Senior Researcher with the Department of Signal Processing, Tampere University of Technology, Tampere, Finland, and an External Researcher with the Nokia Research Center, Tampere. Since 2014, he has been a Faculty Member with Xi’an Jiaotong-Liverpool University, Suzhou, China. His research interests include image and video processing, computer vision, and deep learning.
\end{IEEEbiography}

\begin{IEEEbiography}[{\includegraphics[width=1in,height=1.25in,clip,keepaspectratio]{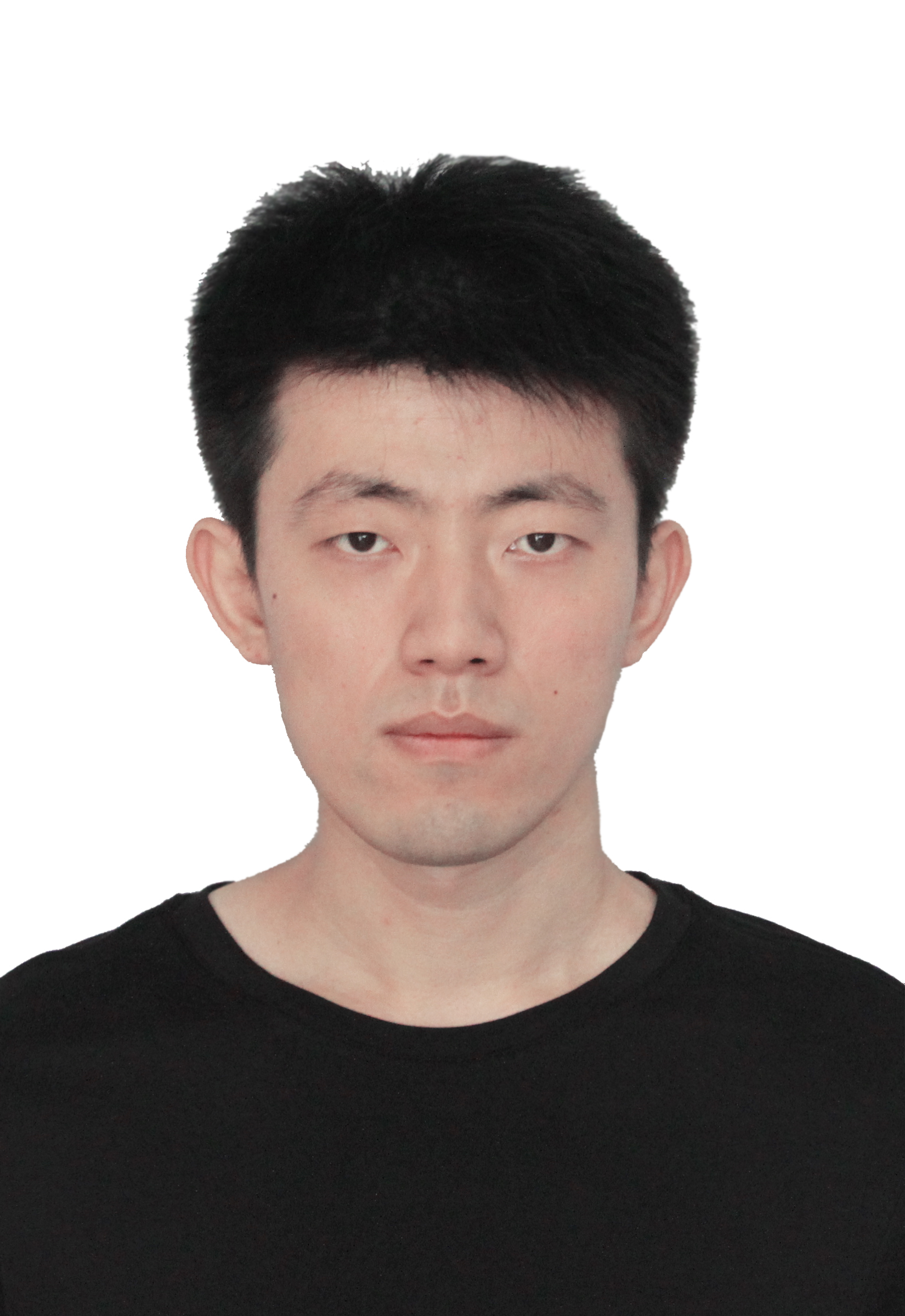}}]{Jianbo Jiao}
	is a Postdoctoral Researcher in the Department of Engineering Science at the University of Oxford. He obtained his Ph.D. degree in Computer Science from City University of Hong Kong in 2018. He was the recipient of the Hong Kong PhD Fellowship. He was a visiting scholar with the Beckman Institute at the University of Illinois at Urbana-Champaign from 2017 to 2018. His research interests include computer vision and machine learning.
\end{IEEEbiography}
\begin{IEEEbiography}[{\includegraphics[width=1in,height=1.25in,clip,keepaspectratio]{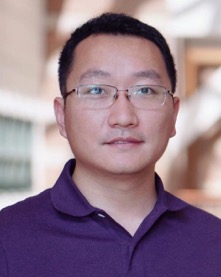}}]{Yunchao Wei}
 is currently an Assistant Professor at the University of Technology Sydney. He received his PhD degree from Beijing Jiaotong University in 2016. Before joining UTS, he was a Postdoc Researcher in Prof. Thomas Huang's Image Formation and Professing (IFP) group at Beckman Institute, UIUC, from 2017 to 2019. He has published over 60 papers on top-tier journals and conferences (e.g., T-PAMI, CVPR, ICCV, etc.), Google citation 3900+. He received the Excellent Doctoral Dissertation Award of CIE in 2016, ARC Discovery Early Career Researcher Award in 2019, 1st Prize in Science and Technology awarded by China Society of Image and Graphics in 2019. His research interests mainly include Deep learning and its applications in computer vision, e.g., image classification, video/image object detection/segmentation, and learning with imperfect data. He has organized multiple Workshops and Tutorials in CVPR, ICCV, ECCV and ACM MM.
\end{IEEEbiography}
\vfill
\begin{IEEEbiography}[{\includegraphics[width=1in,height=1.25in,clip,keepaspectratio]{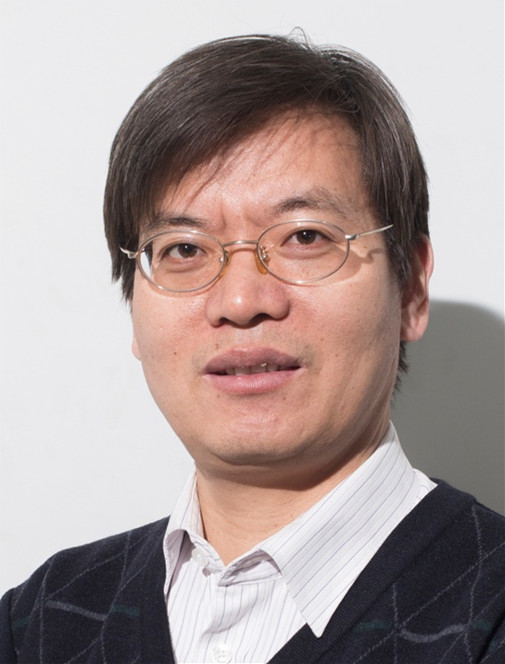}}]{Yao Zhao}
	received the B.S. degree from the Radio Engineering Department, Fuzhou University, Fuzhou, China, in 1989, the M.E. degree from the Radio Engineering Department, Southeast University, Nanjing, China, in 1992, and the Ph.D. degree from the Institute of Information Science, Beijing Jiaotong University (BJTU), Beijing, China, in 1996, where he became an Associate Professor and a Professor in 1998 and 2001, respectively. From 2001 to 2002, he was a Senior Research Fellow with the Information and Communication Theory Group, Faculty of Information Technology and Systems, Delft University of Technology, Delft, The Netherlands. In 2015, he visited the Swiss Federal Institute of Technology, Lausanne (EPFL), Switzerland. From 2017 to 2018, he visited University of Southern California. He is currently the Director with the Institute of Information Science, BJTU. His current research interests include image/video coding, digital watermarking and forensics, video analysis and understanding, and artificial intelligence. Dr. Zhao is a Fellow of the IET. He serves on the Editorial Boards of several international journals, including as an Associate Editor for the IEEE TRANSACTIONS ON CYBERNETICS, a Senior Associate Editor for the IEEE SIGNAL PROCESSING LETTERS, and an Area Editor for Signal Processing: Image Communication. He was named a Distinguished Young Scholar by the National Science Foundation of China in 2010 and was elected as a Chang Jiang Scholar of Ministry of Education of China in 2013. 
\end{IEEEbiography}
\vfill

\end{document}